\documentclass[acmsmall,screen]{acmart}
\AtBeginDocument{%
  }

\setcopyright{acmlicensed}
\copyrightyear{2018}
\acmYear{2018}
\acmDOI{XXXXXXX.XXXXXXX}

\acmJournal{JACM}
\acmVolume{37}
\acmNumber{4}
\acmArticle{111}
\acmMonth{8}

\usepackage{booktabs}
\usepackage{graphicx}
\usepackage{wrapfig}
\newsavebox\CBox

\newcommand{\chunk}[2]{%
\fcolorbox{black}{yellow}{\bfseries\sffamily\scriptsize#1}%
{$\blacktriangleright$#2$\blacktriangleleft$}%
}
\usepackage{svg}

\newcommand{\pablo}[1]{\chunk{Pablo}{\textbf{\textcolor{blue}{\textsl{#1}}}}}

\usepackage{tcolorbox}  
\usepackage{flushend}
\usepackage{orcidlink}
\let\oldnl\nl
\newcommand{\nonl}{\renewcommand{\nl}{\let\nl\oldnl}}
\usepackage{float}
\newtcolorbox{custombox}[1]{
	colback=gray!20,
	colframe=gray!50,
	left=1mm,
	right=1mm,
	top=1mm,
	bottom=1mm,
	fonttitle=\bfseries,
	arc=2mm,
	leftrule=0mm,
	rightrule=.5mm,
	toprule=0mm,
	bottomrule=.5mm,
	notitle,
	before=\par\smallskip\noindent,
	before upper={\textbf{#1: } },
}

\begin{document}

\title{VoxelSBFL: Voxel-Level Spectrum-Based Failure Prediction for Vision-Language-Action Models}

\author{Arusa Kanwal}
\authornote{Both authors contributed equally to this research.}
\email{akanwal@mondragon.edu}
\orcid{0009-0005-4896-8138}
\affiliation{%
  \institution{Mondragon University}
  \city{Mondragon}
  \state{Guipuzcoa}
  \country{Spain}
}

\author{Pablo Valle}
\authornotemark[1]
\email{pvalle@mondragon.edu}
\orcid{0000-0002-0588-316X}
\affiliation{%
  \institution{Mondragon University}
  \city{Mondragon}
  \state{Guipuzcoa}
  \country{Spain}
}

\author{Paola Roggero}
\email{paola.roggero@alumni.mondragon.edu}
\orcid{0009-0000-8124-0492}
\affiliation{%
  \institution{Mondragon University}
  \city{Mondragon}
  \state{Guipuzcoa}
  \country{Spain}
}

\author{Miren Illarramendi}
\email{millarramendi@mondragon.edu}
\orcid{0000-0003-3770-1495}
\affiliation{%
  \institution{Mondragon University}
  \city{Mondragon}
  \state{Guipuzcoa}
  \country{Spain}
}

\author{Shaukat Ali}
\email{shaukat@simula.no}
\orcid{0000-0002-9979-3519}
\affiliation{%
  \institution{Simula Research Laboratory}
  \city{Oslo}
  \country{Norway}
}

\author{Aitor Arrieta}
\email{aarrieta@mondragon.edu}
\orcid{0000-0001-7507-5080}
\correspondingauthor
\affiliation{%
  \institution{Mondragon University}
  \city{Mondragon}
  \state{Guipuzcoa}
  \country{Spain}
}


\begin{abstract}
Vision-Language-Action (VLA) models integrate visual perception, language understanding, and action generation to control robots. Testing these systems produces trajectories and task outcomes, but existing approaches provide limited guidance on using these recorded trajectories to assess subsequent executions or inform policy repair. We conjecture that, for a given policy and task, traversing certain workspace regions is associated with a higher likelihood of failure. To investigate this conjecture, we propose \textit{VoxelSBFL}, a black-box adaptation of Spectrum-Based Fault Localization (SBFL) in which workspace voxels replace program elements as units of analysis. Recorded tool-center-point (TCP) trajectories provide binary voxel coverage, which is combined with execution outcomes to construct suspiciousness maps using established SBFL formulas. These maps identify failure-associated regions and score unseen trajectories without requiring internal model access. We evaluate 34 spectrum-based techniques across four VLA models using 96,000 Pick-Up trajectories in simulation and compare the leading techniques against eight action- and trajectory-based metrics. 
VoxelSBFL obtains an overall macro ROC-AUC of 0.704, significantly outperforming state-of-the art baselines. Sensitivity analyses show that performance varies with spatial granularity and training-set size. 
These findings support spatial spectra for VLA execution assessment and motivate future research on runtime monitoring, self-healing and targeted policy improvement.

\end{abstract}




\maketitle

\section{Introduction}

Vision-Language-Action (VLA) models~\cite{zhong2025survey} integrate visual perception, language understanding, and action generation to control robots that follow users' instructions. Models such as the GR00T-N1 VLA family~\cite{nvidia2025gr00tn1openfoundation} and $\pi_0$~\cite{black2024pi0visionlanguageactionflowmodel} translate visual observations and natural-language instructions into robotic actions. Within the complete robotic stack, these actions are executed by low-level controllers, producing trajectories through the physical workspace. Therefore, assessing the reliability of VLA-enabled robots requires understanding when and under which conditions their executions fail. To support this assessment, recent approaches have proposed different testing methods, including test generation approaches~\cite{wang2025vlatest,kanwal2026fatevla} or methods that aim to deal with the test oracle problem~\cite{valle2026metamorphic,valle2026mango}. 
However, once these tests have been executed and their outcomes determined, existing testing approaches offer limited guidance on how to use the recorded trajectories to predict failures in subsequent executions or inform policy repair. Certain approaches aim at predicting failures during VLA executions. For instance, Valle et al.~\cite{valle2025evaluating} propose both white-box and black-box approaches to distinguish successful and failed executions or high, medium or low quality task executions. SAFE~\cite{gu2026safe} and VLA-Tug~\cite{podolinsky2026failure} rely on internal VLA representations, restricting their use to policies that expose those representations. VLA-Tug also uses annotated failure onsets during training. Nevertheless, these approaches have certain limitations. From the debugging perspective, the scores and alerts produced by these methods give developers limited guidance and explainability of under what circumstances the VLA fails. From the practical perspectives, some of such approaches require internal access to the VLA policy (e.g., white-box uncertainty metrics from~\cite{valle2025evaluating}, internal VLA activations at each policy step for obtaining training data~\cite{podolinsky2026failure}), which is not always easy to obtain.

\begin{wrapfigure}{r}{0.65\linewidth}
  \centering
  \vspace{-0.2cm}
  \includegraphics[trim = 0 30 0 0,width=\linewidth]{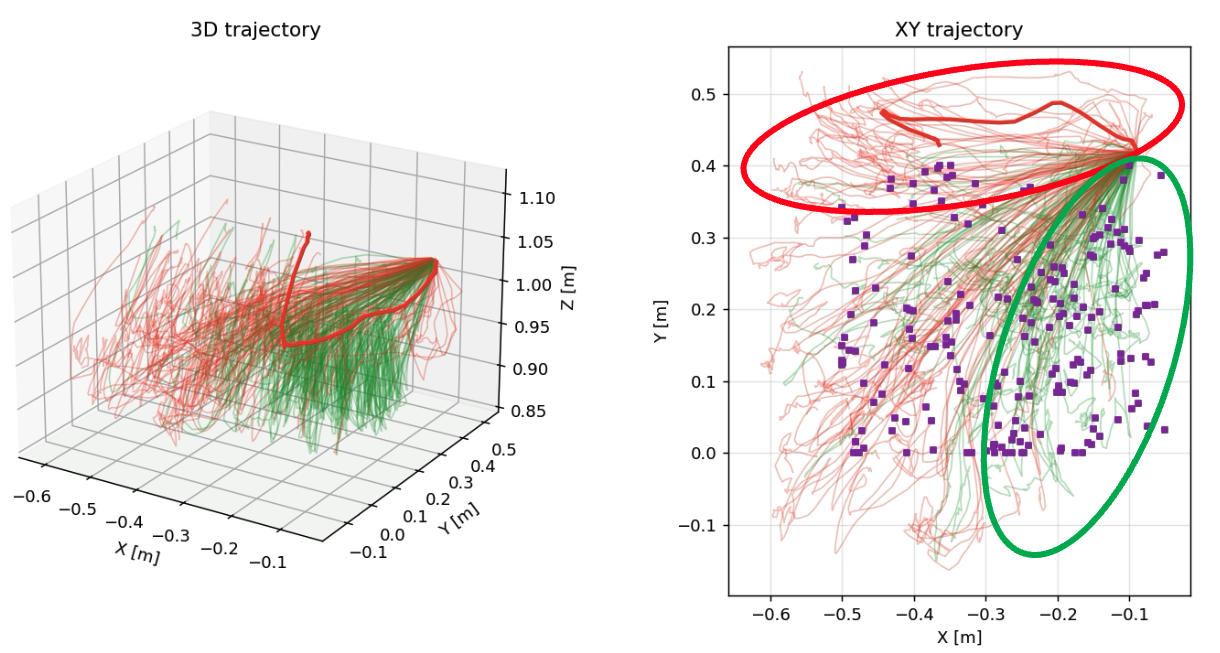}
  \caption{Spatial Distribution of Successful and Failed Robot Trajectories}
  \label{fig:failuredistr}
\end{wrapfigure}

Figure \ref{fig:failuredistr} shows the robot’s end-effector trajectories in three dimensions (left) and projected onto the 2D X-Y plane (right), with failed executions shown in red and successful ones in green for the $\pi_0$ model. The 2D projection makes the spatial pattern particularly clear. Failed trajectories predominate in the upper region outlined in red, while successful trajectories predominate in the right-hand region outlined in green. We therefore conjecture that, in VLA-enabled robots, traversing certain regions of the workspace is associated with a higher likelihood of task failure. This conjecture draws on an analogy with conventional software. A test that executes a faulty statement may be more likely to fail because certain inputs or program states can trigger the fault. Similarly, a robot execution may be more likely to fail when the end effector passes through a problematic region of the workspace. This analogy motivates treating the regions visited by a robot as a form of coverage, much like the statements executed by a software test.

Building on this intuition, we propose VoxelSBFL, which recasts failure prediction for VLA-enabled robots as a spatial adaptation of spectrum-based fault localization (SBFL)~\cite{jones2002visualization}. We partition the robot’s three-dimensional workspace into voxels and use these voxels in place of program elements. Just as a software test covers a statement by executing it, a robot trajectory covers a voxel if at least one recorded position of the end effector’s tool center point falls within it. We combine this voxel coverage information with the success or failure outcome of each execution and apply standard SBFL formulas to quantify how strongly coverage of each voxel is associated with failure. 
The resulting suspiciousness map highlights workspace regions that warrant closer inspection. Developers can use it to identify segments of failed trajectories to examine, including the actions and observations that preceded entry into a highly suspicious voxel. The same map could guide the generation of new trajectories that exercise these regions and the collection of successful demonstrations for targeted repair (e.g., through policy fine-tuning). In this study, we evaluate how well the map distinguishes successful from failed executions, as a proof to determine whether spectrum-based techniques can be employed in this context.

We evaluate how well \textit{VoxelSBFL} distinguishes successful from failed executions using complete, held-out trajectories. We compare 34 SBFL formulas on 96,000 simulated executions across four state-of-the-art VLA models and benchmark the leading formulas against eight other black-box metrics proposed by Valle et al. \cite{valle2025evaluating}, including action- and trajectory-based metrics. Harmonic Mean and Rogot2 achieve the highest macro-averaged ROC-AUC among the SBFL formulas (0.704), outperforming the strongest motion-instability baseline (0.655). We also find that voxel size affects the relative ranking of the formulas and that Harmonic Mean improves as more labeled trajectories are used to construct the map. These results not only suggest that VoxelSBFL can predict failing executions, but that certain VLA failures may be associated with certain spatial regions and that certain spectrum-based techniques can help in identifying suspicious voxels.

The key contributions of this paper can be summarized as follows:

\begin{itemize}
    \item We introduce VoxelSBFL, a black-box adaptation of spectrum-based fault localization that uses end-effector visits to workspace voxels as coverage and combines them with episode outcomes. 
    \item Through a large-scale evaluation across four VLA models, we show that voxel coverage can distinguish successful from failed trajectories and that the leading SBFL formulas outperform generic action- and motion-instability metrics overall.
    \item  We characterize how voxel size and training-set size affect performance and document the drop in discrimination when transferring from Pick-Up to Move-Near. 
    \item We show how voxel suspiciousness maps can help developers prioritize trajectory segments for inspection and discuss their potential for targeted test generation, policy repair, and runtime monitoring.
    \item We provide a replication package~\cite{kanwal2026voxelsbfl_replication} to support replication of the experiments and extension of the evaluation to other VLA models and robotic tasks by other researchers.
\end{itemize}

The rest of the paper is structured as follows. Section~\ref{sec:background} provides background on Vision-Language-Action (VLA) models and Spectrum-Based Fault Localization (SBFL). Section~\ref{sec:approach} presents the proposed VoxelSBFL approach, including workspace voxelization, trajectory coverage, and voxel suspiciousness computation. Section~\ref{sec:EmpiricalEvaluation} describes the empirical evaluation setup and research questions. Section~\ref{sec:ExpResult} presents and discusses the experimental results, including comparisons of SBFL techniques, voxel-size and coverage sensitivity, and cross-task generalizability. Section~\ref{sec:ThreatstoValidity} discusses threats to validity, while Section~\ref{sec:Relatedwork} reviews related work. Finally, Section~\ref{sec:conclusion}  concludes the paper and outlines directions for future research.

\section{Background}
\label{sec:background}

\subsection{Vision-Language-Action Models}

Vision-Language-Action models (VLAs) are multi-modal models that integrate visual perception, language understanding, and action generation to control robots~\cite{nvidia2025gr00tn1openfoundation, black2024pi0visionlanguageactionflowmodel, kim2024openvla, qu2025spatialvla, qu2025eo1}. Robotic systems that use these models for decision-making constitute an emerging class of Cyber-Physical Systems (CPSs), in which learned policies interconnect sensor inputs and user instructions (in the form of natural language) with physical actions. While conventional robotic architectures typically organize perception, planning, and control into separate components, VLA models learn a direct mapping from multi-modal observations into actions. Their architectures combine visual and language representations along with proprioceptive information to generate the next action(s) to be taken. This enables a robot to interpret a natural-language-based instruction such as ``\textit{Pick up the apple}'' in the context of its environment, identify the intended object, and generate the movements required to interact with it and follow the instruction given by the human.

\begin{figure*}[ht]
    \centering
    \includegraphics[width=\textwidth]{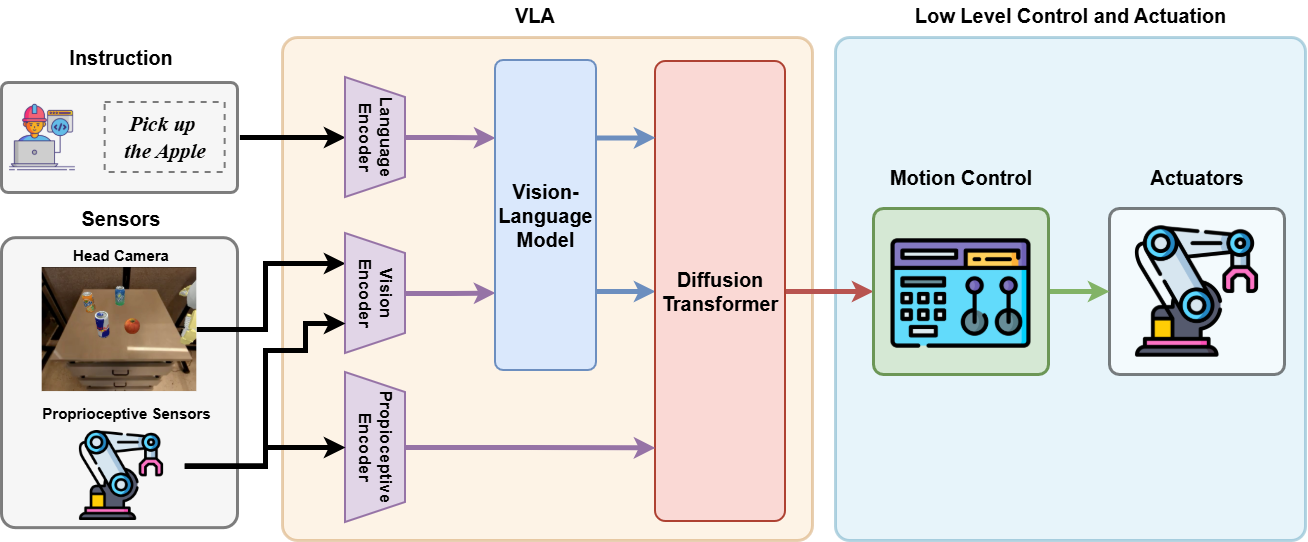}
    \caption{Overview of the Architecture of a VLA-Enabled Robot}
    \label{fig:VLA}
\end{figure*}

Figure~\ref{fig:VLA} shows the interaction between the sensors, VLA-based decision-making, and low-level control in a VLA-enabled robot. Sensors capture the visual observations and the robot's proprioceptive state, which are provided to the VLA model along with the user instruction. Within the model, modality-specific encoders and multi-modal processing components produce contextual representations used by an action decoder to generate the control commands. These commands are subsequently passed to the robot's low-level control and actuation components, which translate them into platform-specific actuator commands employing traditional well-known controllers (e.g., Proportional-Integral-Derivative (PID) controllers). 

Formally, at a decision step $t$, the model receives an observation, $o_t=\left[I^t_1, I^t_2, \dots, I^t_n, \ell_t, \theta_t\right]$, where $I^t_i$ denotes an RGB image captured by camera $i$, $\ell_t$ is the natural-language instruction, and $\theta_t$ represents the robot's proprioceptive state, such as joint positions and velocities, or gripper forces. For models that use action chunking~\cite{black2024pi0visionlanguageactionflowmodel}, the policy models a conditional distribution $p(AC_t \mid o_t)$ over action chunks, $AC_t =
\left[a_t, a_{t+1}, \dots, a_{t+H-1}\right]$, where $H$ is the prediction horizon and each $a_{t+j} \in \mathbb{R}^{D}$ represents a control command with $D$ dimensions. The action representation depends on the robot and policy as it may specify joint targets, end-effector poses or displacements, and gripper commands. The prediction horizon also varies across models and configurations (e.g., $\pi_0$ employed chunks of $H=50$ size in their original paper~\cite{black2024pi0visionlanguageactionflowmodel}). 

By iteratively generating action chunks, the VLA model produces control commands that, when executed by the robot, induce a trajectory. Let $AC_{t_i}$ denote the action chunk generated at decision step $t_i$, and let $E_i$ denote the sequence of commands actually executed before the next chunk is generated. Depending on the execution strategy, $E_i$ may contain all actions in $AC_{t_i}$ or only a subset, for example when a new prediction replaces the current chunk before its execution is complete since the execution horizon is shorter than the generation horizon. The resulting sequence of executed commands is $A = \left[E_1, E_2, \dots, E_N\right]$, where $N$ is the number of applied chunks. Executing these commands produces a trajectory $\mathcal{T} = \left[S_t, S_{t+1}, \dots, S_{t+T}\right]$, where $T = \sum_{i=1}^{N} |E_i|$ is the total number of executed actions. For manipulation tasks, each state $S_t$ can be characterized by the configuration of the robot's end-effector, i.e., Tool Center Point (TCP), including its position $(x,y,z)$ and orientation, represented by a unit quaternion $(q_1,q_2,q_3,q_4)$. The robot's low-level control system translates the generated actions into joint movements, allowing the VLA model to connect natural-language instructions and visual observations with physically grounded behavior.

VLA models differ in how they generate actions. While some architectures predict discretized action tokens autoregressively~\cite{kim2024openvla}, models such as $\pi_0$~\cite{black2024pi0visionlanguageactionflowmodel} and the GR00T models (i.e., Gr00t-N1.5, Gr00t-N1.6, and Gr00t-N1.7)~\cite{nvidia2025gr00tn1openfoundation} use conditional flow matching to generate continuous action chunks. At inference time, these models initialize an action chunk with noise and iteratively transform it into an action sample conditioned on visual, linguistic, and proprioceptive information. In GR00T models, this process is implemented using a diffusion transformer that combines state and action representations with features from a vision-language backbone. These iterative updates are internal steps of action generation and are distinct from the physical control steps at which the robot executes actions. VLA models are commonly trained on robot demonstration datasets that associate visual observations and natural-language instructions with the actions performed by a robot~\cite{kim2024openvla, black2024pi0visionlanguageactionflowmodel, nvidia2025gr00tn1openfoundation}. From these demonstrations, the model learns to relate instructions to the observed environment and generate actions that accomplish the requested task. Large-scale datasets, such as Open X-Embodiment~\cite{o2024open}, provide demonstrations spanning multiple tasks and robot embodiments, supporting the learning of transferable visuomotor behaviors. Subsequent fine-tuning and task- or robot-specific demonstrations adapt the policy to the target deployment setting, including its observation and action interfaces~\cite{kim2024openvla}. This adaptation can improve task performance, but does not by itself guarantee that generated actions satisfy the robot's physical constraints or the system's safety requirements.

\subsection{Spectrum-based Fault Localization}

\begin{table*}[t]
\centering
\caption{SBFL example using Tarantula. A bullet ($\bullet$) denotes execution and a dash (--) denotes non-execution; F and P denote failing and passing tests, respectively. The faulty block and its score are highlighted in bold.}
\label{tab:sbfl-example}
\small
\setlength{\tabcolsep}{5pt}
\renewcommand{\arraystretch}{1.15}
\begin{tabular}{@{}cl*{5}{c}rrrrrc@{}}
\hline
\textbf{ID} & \textbf{Program block}
& $t_1$ & $t_2$ & $t_3$ & $t_4$ & $t_5$
& $N_{CF}$ & $N_{CS}$ & $N_F$ & $N_S$
& \textbf{Suspiciousness} & \textbf{Rank}\\
\hline

$b_1$ &
\shortstack[l]{\texttt{fee = 0}\\
               \texttt{if priority:}}
& $\bullet$ & $\bullet$ & $\bullet$ & $\bullet$ & $\bullet$
& 2 & 3 & 2 & 3 & 0.500 & 3\\

$b_2$ & \quad\texttt{fee += 5}
& $\bullet$ & -- & $\bullet$ & $\bullet$ & --
& 1 & 2 & 2 & 3 & 0.429 & 6\\

$b_3$ & \texttt{if amount\_based:}
& $\bullet$ & $\bullet$ & $\bullet$ & $\bullet$ & $\bullet$
& 2 & 3 & 2 & 3 & 0.500 & 3\\

$b_4$ & \quad\texttt{if amount > 200:}
& $\bullet$ & $\bullet$ & $\bullet$ & -- & --
& 2 & 1 & 2 & 3 & 0.750 & 2\\

$\mathbf{b_5}$ & \qquad\textbf{\texttt{fee += 0.20 * amount}}
& $\bullet$ & $\bullet$ & -- & -- & --
& 2 & 0 & 2 & 3 & \textbf{1.000} & \textbf{1}\\

$b_6$ &
\shortstack[l]{\quad\texttt{else:}\\
               \qquad\texttt{fee += 0.10 * amount}}
& -- & -- & $\bullet$ & -- & --
& 0 & 1 & 2 & 3 & 0.000 & 7\\

$b_7$ &
\shortstack[l]{\texttt{else:}\\
               \quad\texttt{fee += 10}}
& -- & -- & -- & $\bullet$ & $\bullet$
& 0 & 2 & 2 & 3 & 0.000 & 7\\

$b_8$ & \texttt{return fee}
& $\bullet$ & $\bullet$ & $\bullet$ & $\bullet$ & $\bullet$
& 2 & 3 & 2 & 3 & 0.500 & 3\\

\hline
\multicolumn{2}{@{}l}{\textbf{Test outcome}}
& \textbf{F} & \textbf{F} & \textbf{P} & \textbf{P} & \textbf{P}
& \multicolumn{6}{c@{}}{}\\
\hline
\end{tabular}
\end{table*}

Spectrum-based fault localization (SBFL) assists debugging by ranking program components according to their association with observed test failures~\cite{jones2002visualization,abreu2009practical,wong2016survey}. It combines execution information, represented as \emph{program spectra}, with test outcomes. For example, a binary \emph{block-hit spectrum} records whether each code block is executed at least once by a test. Using these spectra and the corresponding pass or fail outcomes, SBFL assigns each component a \emph{suspiciousness score} and ranks components in descending order of suspiciousness to guide inspection. Different SBFL techniques use different formulas to calculate suspiciousness~\cite{naish2011model,wong2016survey}. A representative technique is \emph{Tarantula}~\cite{jones2002visualization}. For a covered block $b$, assuming at least one failing and one passing test, its score is:

\begin{equation}
\operatorname{Susp}_{\mathrm{Tarantula}}(b) =
\frac{N_{CF}(b)/N_F}
     {N_{CF}(b)/N_F + N_{CS}(b)/N_S},
\label{eq:sbfl-tarantula}
\end{equation}
where $N_{CF}(b)$ and $N_{CS}(b)$ denote the numbers of failing and passing tests that execute $b$, respectively, and $N_F$ and $N_S$ denote the total numbers of failing and passing tests. Tarantula therefore assigns higher scores to blocks executed by a larger proportion of failing tests and a smaller proportion of passing tests.

Table~\ref{tab:sbfl-example} illustrates this using a buggy implementation of a fee-calculation program and five tests, of which two fail and three pass ($N_F=2$, $N_S=3$). The bug lies in block $b_5$, where the multiplier is $0.20$ instead of the intended $0.10$. Both failing test cases execute $b_5$, whereas no passing test case executes it. Consequently, Tarantula assigns this block the maximum score of $1.000$, placing it first in the inspection ranking. In this example, a block's rank is one plus the number of blocks with a strictly higher score; blocks with identical scores therefore share the same rank.


\section{Approach}
\label{sec:approach}

\begin{figure*}[ht]
    \centering
    \includegraphics[trim = 0 50 0 0,width=\linewidth]{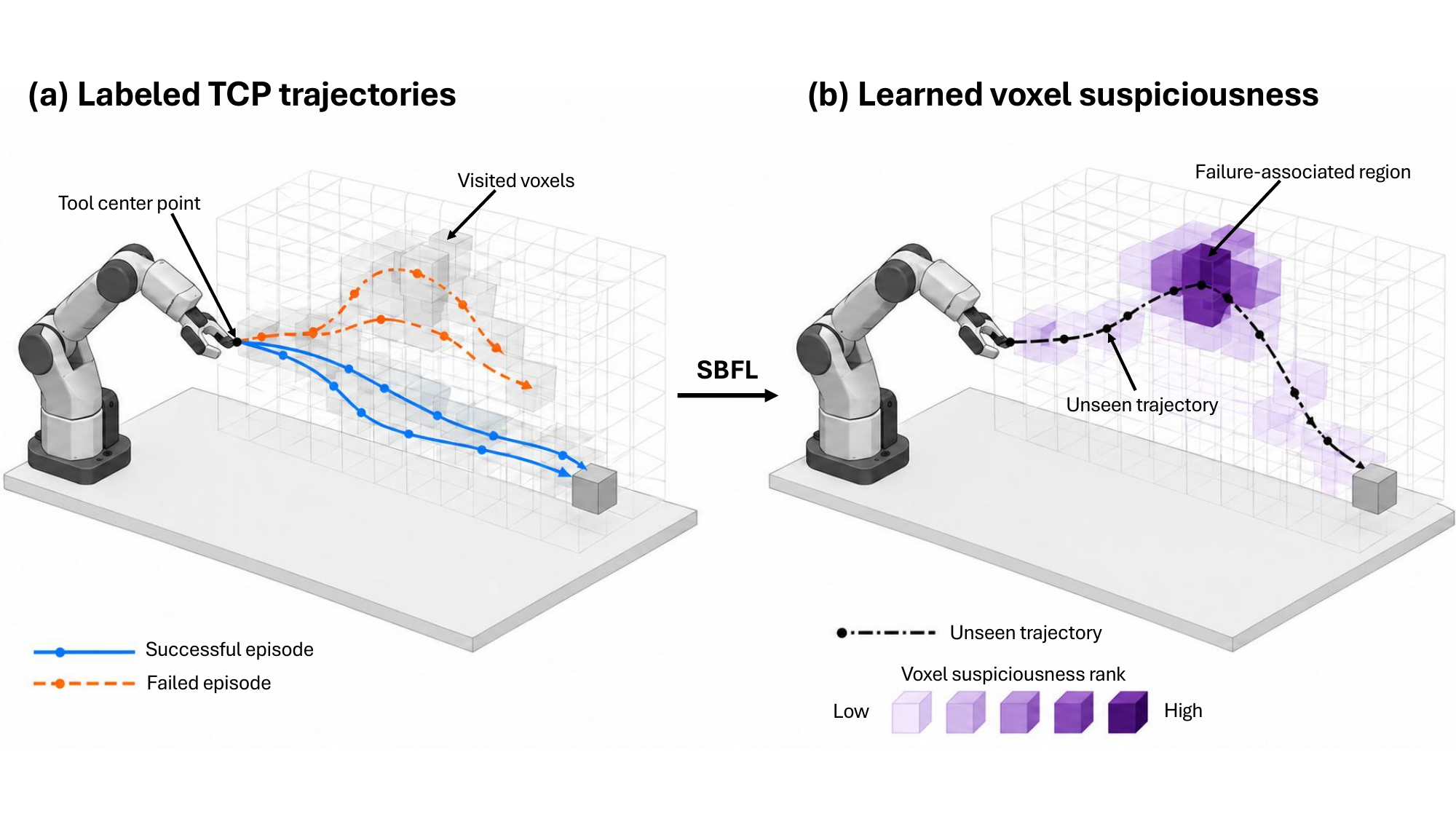}
    \caption{Overview of VoxelSBFL. Successful and failed
    tool-center-point trajectories are mapped onto a voxel grid.
    SBFL ranks visited voxels by their association with failed
    executions. The resulting suspiciousness map is used to
    score unseen trajectories and predict whether they are going to fail.}
    \label{fig:voxelsbfl-overview}
\end{figure*}

We propose a spatial adaptation of Spectrum-Based Fault Localization (SBFL) for VLAs. Unlike in traditional SBFL \cite{wong2016survey}, which uses program elements, such as statements or methods, our approach uses spatial regions of the robot workspace. 
Robot trajectories then play the role of test executions: a trajectory covers a spatial region when the robot visits it, and its task outcome determines whether the execution passes or fails.
The approach requires only recorded robot positions and episode-level outcomes. A key advantage of our approach is that it does not require access to the internal parameters, gradients, activations, or architecture of the VLA model. Therefore, unlike other approaches, such as those aiming to measure uncertainty \cite{zollo2025confidence,karli2025ask,valle2025evaluating,chen2026reconvla}, our method can be applied as a non-invasive, model-agnostic analysis procedure to determine at run-time when a failure is about to happen. The approach consists of four main stages: (1) voxelizing the robot workspace; (2) transforming labeled trajectories into binary voxel-coverage spectra; (3) computing voxel suspiciousness; and (4) applying the resulting suspiciousness map to unseen trajectories. Figure \ref{fig:voxelsbfl-overview} depicts an intuitive representation of VoxelSBFL and an overview of the approach.

\subsection{Workspace Voxelization}
\label{sec:voxelization}

Let the robot workspace of a VLA-enabled robot be a three-dimensional region $\mathcal{W}\subseteq\mathbb{R}^{3}$. We partition this workspace using a regular voxel grid defined by an origin $\mathbf{o}=(o_x,o_y,o_z)$, and a voxel size $s>0$. Given a Cartesian robot position $\mathbf{p}=(p_x,p_y,p_z)\in\mathcal{W}$, its voxel index is obtained as

\[
    \nu_s(\mathbf{p})
    =
    \left\lfloor
    \frac{\mathbf{p}-\mathbf{o}}{s}
    \right\rfloor
    \in \mathbb{Z}^{3},
\]

\noindent where subtraction, division, and the floor operation are applied element-wise. A voxel with index $\mathbf{v}=(v_x,v_y,v_z)$ represents the half-open spatial region, formalized as follows:

\[
    \mathcal{V}_{\mathbf{v}}
    =
    \prod_{d\in\{x,y,z\}}
    \left[
        o_d+s v_d,\,
        o_d+s(v_d+1)
    \right).
\]

This definition ensures that every position is assigned to exactly one voxel, including positions lying on grid boundaries. Although the grid is conceptually unbounded, the implementation stores only voxels visited by at least one training trajectory.
The voxel size controls the spatial granularity of the analysis. Smaller voxels provide finer localization but produce a larger and sparser spectrum. Larger voxels combine nearby positions, increasing support but reducing spatial resolution.\footnote{This trade-off is examined in our experiments covering RQ3.}

\subsection{Trajectory and Outcome Representation}
\label{sec:trajectory-representation}

A robot episode $i$ produces a time-ordered trajectory, i.e., $\tau_i =
    \left(
        \mathbf{p}_{i,1},
        \mathbf{p}_{i,2},
        \ldots,
        \mathbf{p}_{i,T_i}
    \right)$,
where $\mathbf{p}_{i,t}\in\mathbb{R}^{3}$ is the recorded Cartesian position at time step $t$, and $T_i$ is the number of recorded steps.
Each episode is accompanied by an externally provided binary outcome $y_i$, where $y_i=1$ refers to episode failure, while $y_i=0$ refers to episode pass.
The outcome is not inferred from the trajectory and is not produced by the SBFL approach. It is determined by the success criterion of the underlying manipulation task.

The voxel coverage set of trajectory $\tau_i$ is
$\mathcal{C}_i
    =
    \left\{
        \nu_s(\mathbf{p}_{i,t})
        \mid
        1\leq t\leq T_i
    \right\}$.
Coverage is binary at the trajectory level. We define
$x_{i,\mathbf{v}}
    =
    \mathbb{I}
    \left[
        \mathbf{v}\in\mathcal{C}_i
    \right]$,
\noindent where $\mathbb{I}[\cdot]$ is the indicator function. Therefore, a trajectory contributes at most one coverage observation to each voxel, regardless of how many times it visits that voxel. This representation follows the semantics of conventional SBFL, in which a program element is either executed or not executed by a test case. It also prevents long trajectories or trajectories that remain in one spatial region for many time steps from dominating the spectrum. The current formulation consequently captures spatial visitation but does not use aspects such as visit frequency, time spent in a voxel, or temporal ordering.

\subsection{Construction of the Voxel Spectrum}
\label{sec:voxel-spectrum}

Let $\mathcal{D}_{\mathrm{train}}$ denote a collection of labeled training trajectories. The set of training-observed voxels is
$\mathcal{V}_{\mathrm{train}}
    =
    \bigcup_{i\in\mathcal{D}_{\mathrm{train}}}
    \mathcal{C}_i$.
For every voxel
$\mathbf{v}\in\mathcal{V}_{\mathrm{train}}$, we construct a four-element coverage spectrum: 

\[
\begin{aligned}
    e_f(\mathbf{v})
    &=
    \sum_{i\in\mathcal{D}_{\mathrm{train}}}
    y_i x_{i,\mathbf{v}},\\
    e_p(\mathbf{v})
    &=
    \sum_{i\in\mathcal{D}_{\mathrm{train}}}
    (1-y_i)x_{i,\mathbf{v}},\\
    n_f(\mathbf{v})
    &=
    \sum_{i\in\mathcal{D}_{\mathrm{train}}}
    y_i(1-x_{i,\mathbf{v}}),\\
    n_p(\mathbf{v})
    &=
    \sum_{i\in\mathcal{D}_{\mathrm{train}}}
    (1-y_i)(1-x_{i,\mathbf{v}}).
\end{aligned}
\]

\noindent where $e_f(\mathbf{v})$ refers to failed  trajectories that visit voxel $\mathbf{v}$; $e_p(\mathbf{v})$ refers to successful trajectories that visit voxel $\mathbf{v}$; $n_f(\mathbf{v})$ refers to failed trajectories that do not visit voxel $\mathbf{v}$; and$n_p(\mathbf{v})$: successful trajectories that do not visit voxel $\mathbf{v}$.
Every training trajectory contributes exactly once to one of these four categories for each voxel. Therefore, $e_f(\mathbf{v})+n_f(\mathbf{v})=N_f$
and $e_p(\mathbf{v})+n_p(\mathbf{v})=N_p$, where $N_f$ and $N_p$ are the total numbers of failed and successful training trajectories, respectively.

A voxel with a large $e_f$ and a small $e_p$ is visited more frequently by failed trajectories than by successful ones and may consequently be associated with failure. However, the precise relationship between these four counts is determined by the selected SBFL technique. Voxels that are never observed during training are not assigned a spectrum or suspiciousness value. Assigning them a score based on $e_f=e_p=0$ would incorrectly imply that the training data contain evidence about their behavior.

Voxel visitation also provides a basis for assessing the \textit{spatial adequacy} of a test suite, which we call the \textbf{Workspace Voxel Coverage (WVC)}. Let $\mathcal{T}$ be a suite of test trajectories, let $\mathcal{C}_j$ denote the set of voxels visited by trajectory $\tau_j$, and let $\mathcal{V}_{\mathrm{ref}}$ be a fixed, nonempty set of reference voxels representing the workspace where the root operates. We define the \emph{WVC} of $\mathcal{T}$ as the proportion of reference voxels visited by at least one trajectory in the suite:
\begin{equation}
    \operatorname{WVC}(\mathcal{T}; \mathcal{V}_{\mathrm{ref}})
    =
    \frac{
        \left|
            \mathcal{V}_{\mathrm{ref}} \cap
            \bigcup_{\tau_j \in \mathcal{T}} \mathcal{C}_j
        \right|
    }{
        \left|\mathcal{V}_{\mathrm{ref}}\right|
    }.
    \label{eq:workspace-voxel-coverage}
\end{equation}
A test suite satisfies the workspace voxel coverage criterion at a
prespecified threshold $\theta \in (0,1]$ if $\operatorname{WVC}(\mathcal{T}; \mathcal{V}_{\mathrm{ref}}) \geq \theta$. For example, a threshold of $0.8$ requires the suite to visit at least 80\% of the reference voxels. Comparisons between test suites shall use the same coordinate frame, voxel grid, and reference set.

\subsection{Voxel Suspiciousness}
\label{sec:voxel-suspiciousness}

Let $g_m$ denote the suspiciousness function associated with the SBFL technique $m$ (e.g., Trantula, Ochiai). The raw suspiciousness of voxel $\mathbf{v}$ is $\sigma_m(\mathbf{v})
    =
    g_m\left(
        e_f(\mathbf{v}),
        e_p(\mathbf{v}),
        n_f(\mathbf{v}),
        n_p(\mathbf{v})
    \right)$.
Our approach can use any SBFL techniques from the literature. These techniques combine the spectrum counts differently and may react differently to class imbalance, sparse voxel support, and increasing workspace coverage. 
Some SBFL techniques' formulas may be undefined for particular spectra, e.g., when their denominator is zero. We handle such cases by using a predefined finite policy. An undefined $0/0$ expression receives the technique's neutral value, whereas mathematical positive or negative infinities are replaced by finite rank-preserving extreme values. These cases are recorded as diagnostics rather than silently removing the corresponding voxels.

Because the selected techniques produce scores with different numerical ranges, their raw suspiciousness values are transformed into empirical percentile ranks. Let $N_{\mathcal{V}}
    =
    \left|\mathcal{V}_{\mathrm{train}}\right|$
be the number of training-observed voxels. For a voxel $\mathbf{v}$, let
$L_m(\mathbf{v})$ be the number of training voxels with a strictly smaller score than $\sigma_m(\mathbf{v})$, and let $E_m(\mathbf{v})$ be the number with the same score. Its tie-aware percentile is

\[
    r_m(\mathbf{v})
    =
    \frac{
        L_m(\mathbf{v})
        +
        \frac{E_m(\mathbf{v})-1}{2}
    }{
        N_{\mathcal{V}}-1
    },
\]

\noindent when $N_{\mathcal{V}}>1$. If only one voxel is observed during training, it receives the neutral percentile $0.5$.

The percentile $r_m(\mathbf{v})\in[0,1]$ expresses the relative position of the voxel within the suspiciousness map. A value close to one indicates that the voxel is ranked among the most failure-associated spatial regions according to technique $m$. Voxels with identical raw suspiciousness receive the same average percentile.
The output of the training stage is therefore a technique-specific suspiciousness map, i.e., 
$\mathcal{M}_m
    =
    \left\{
        \left(
            \mathbf{v},
            \sigma_m(\mathbf{v}),
            r_m(\mathbf{v})
        \right)
        \mid
        \mathbf{v}\in\mathcal{V}_{\mathrm{train}}
    \right\}$.
Sorting $\mathcal{M}_m$ in descending order of $r_m$ produces a ranked list of failure-associated spatial regions. Because every voxel index can be transformed back into its physical bounds or center position, this ranking can be visualized in the robot workspace and used to identify candidate regions for further inspection or intervention.

\subsection{Application to Unseen Trajectories}
\label{sec:trajectory-scoring}

The suspiciousness map provides the voxel-level localization output. To evaluate whether this map captures spatial behavior associated with failures, we additionally use it to score unseen trajectories.
For an unseen trajectory $\tau_j$, the set of visited voxels for which the training process produced a suspiciousness value is
$ \mathcal{K}_j
    =
    \mathcal{C}_j
    \cap
    \mathcal{V}_{\mathrm{train}}$.
The trajectory-level risk assigned by technique $m$ is

\[
    S_m(\tau_j)
    =
    \begin{cases}
        \displaystyle
        \max_{\mathbf{v}\in\mathcal{K}_j}
        r_m(\mathbf{v}),
        &
        \text{if }\mathcal{K}_j\neq\emptyset,\\[8pt]
        0.5,
        &
        \text{if }\mathcal{K}_j=\emptyset.
    \end{cases}
\]

The maximum operator reflects the hypothesis that visiting one highly suspicious spatial region may be sufficient to place an execution at risk. It also preserves a direct connection between trajectory-level predictions and individual suspicious voxels.
Unknown voxels are ignored when at least one known voxel exists. When a trajectory contains no training-observed voxel, the available data provide no evidence for classifying it as either successful or failed. We therefore assign it the neutral score $0.5$. 
The test trajectory outcome is used only after $S_m(\tau_j)$ has been computed. It is never used to modify the voxel map or select its suspicious regions. This separation allows the failure-discrimination ability of the learned spatial map to be evaluated on previously unseen executions without label leakage.

\section{Empirical Evaluation}
\label{sec:EmpiricalEvaluation}

Our empirical evaluation explores the following five Research Questions (RQs):

\begin{itemize}
    \item \textbf{RQ1 -- Overall SBFL Technique Performance:} \textit{Which spectrum-based technique is the most effective at identifying failure-associated voxels in VLA models?} With this RQ, we aim to determine which SBFL techniques produce suspiciousness maps that best distinguish failed trajectories from successful ones. To this end, we compare the scores derived from each technique under the same experimental conditions.

    \item \textbf{RQ2 -- Comparison baseline:} \textit{How do the best spectrum-based techniques compare with respect to state-of-the-art VLA black-box uncertainty detection methods?} 
    In this RQ, we aim to assess whether the best-performing SBFL techniques can discriminate failures as effectively as existing black-box metrics from a prior study \cite{valle2025evaluating}. To this end, we compare their failure-discrimination performance on the same evaluation data.

    \item \textbf{RQ3 -- Voxel-Size Sensitivity:} \textit{How does voxel size affect localization effectiveness, spatial coverage, computational cost, and the relative performance of spectrum-based techniques?} With this RQ, we aim to understand the sensitivity that voxel size has on the effectiveness of SBFL techniques.

    \item \textbf{RQ4 -- Voxel-Coverage Sensitivity:} \textit{How does the \emph{workspace voxel coverage} achieved by the training trajectories affect the effectiveness and robustness of different SBFL techniques?} In this RQ, we investigate whether the techniques remain effective when the available training trajectories cover only part of the workspace. To this end, we vary the number of training trajectories and examine the resulting voxel coverage and failure-discrimination performance.

    \item \textbf{RQ5 -- Cross-Task Generalizability:} \textit{To what extent do suspiciousness maps learned from one robotic task retain their failure-discrimination effectiveness when applied to a different task?} With this RQ, we aim to determine whether failure-associated spatial patterns learned from Pick-Up trajectories remain useful for identifying failures in a different task. To this end, we evaluate the learned maps from Pick-Up on Move-Near trajectories without retraining or target-task calibration.
\end{itemize}


We evaluate the proposed approach through a sequence of controlled experiments designed to isolate the effects of the SBFL technique, voxel granularity, training coverage, and task distribution. 

RQ1 uses deterministic leave-one-run-out validation within each combination of VLA model, trajectory-generation condition, and scene budget. Nine complete runs are used for training and the remaining run is used for testing, until every run has served as the test set. RQ2 retains the same evaluation partitions for the baseline comparison, while RQ3 uses these partitions to examine the effect of voxel size. RQ4 introduces a leakage-safe three-way partition consisting of training-pool, calibration, and test runs, allowing training subsets with different spatial coverage to be selected without consulting the test data.

All comparisons between techniques are paired: techniques are evaluated using the same trajectories, voxel grids, and experimental partitions. Results are first averaged over repeated runs and subsequently macro-averaged over the experimental configurations, preventing configurations containing more trajectories from dominating the results.

\subsection{Selected SBFL Techniques}

We evaluated 34 spectrum-based techniques reported by Wong et al.~\cite{wong2016survey}. Their corresponding
formulas are described in the survey, while the complete set of techniques evaluated in our study is reported in Table~\ref{tab:rq1-cross-vla}.
Each technique received the same voxel-level spectrum and is evaluated on the same held-out trajectories. No technique-specific parameter tuning or trajectory-aggregation rule was used. Undefined expressions, such as $0/0$, were assigned a predefined neutral value. Mathematical infinities were replaced by finite, rank-preserving values. These cases were recorded separately as diagnostics, ensuring that every technique produces finite scores without silently discarding voxels or trajectories. The complete catalogue was retained in RQ1--RQ4. Consequently, changes in technique ranking can be attributed to the experimental factor under study rather than to differences in the set of techniques being compared.

\subsection{Selected VLA Models}

The evaluation includes four VLA models: \texttt{EO1}~\cite{qu2025eo1}, \texttt{GR00T-N1.6}~\cite{nvidia2025gr00tn1openfoundation}, \texttt{Pi0}~\cite{black2024pi0visionlanguageactionflowmodel}, and \texttt{SpatialVLA}~\cite{qu2025spatialvla}. These models were selected to cover different architectural designs and action generation mechanisms. Specifically, \texttt{EO1} integrates multimodal reasoning and action generation within a shared transformer backbone, whereas \texttt{Pi0} and \texttt{GR00T-N1.6} combine vision-language backbones with dedicated action generation components based on flow matching. In contrast, \texttt{SpatialVLA} incorporates explicit spatial representations and adaptive discretized action grids. Furthermore, these models have been used across multiple empirical studies on robotic manipulation, robustness assessment, and automated testing~\cite{valle2025evaluating, valle2026metamorphic, kanwal2026fatevla, chen2026pointact, vuong2026world2act, wang2025vlatest, peng2025nebula}, supporting their relevance as subjects for our evaluation. Moreover, all of these models are compatible with the open-source simulation environment SimplerEnv~\cite{li24simpler}, allowing our evaluation to be easily replicated by other researchers. 

We conducted our experiments using the environment adopted from state-of-the-art works~\cite{valle2026metamorphic, kanwal2026fatevla, valle2025evaluating, wang2025vlatest}, employing the fine-tuned versions of each model adapted to the fractal benchmark dataset~\cite{brohan2022rt} corresponding to the task categories used in our evaluation suite. By considering multiple VLA policies, we aim to investigate whether the behavior and relative effectiveness of the SBFL techniques remain consistent across models with different architectures and action representations.

\subsection{Simulator, Robot Embodiment and Robotic Tasks}

\begin{figure*}[ht]
    \centering
    \begin{minipage}[t]{0.49\textwidth}
        \centering
        \includegraphics[width=\linewidth]{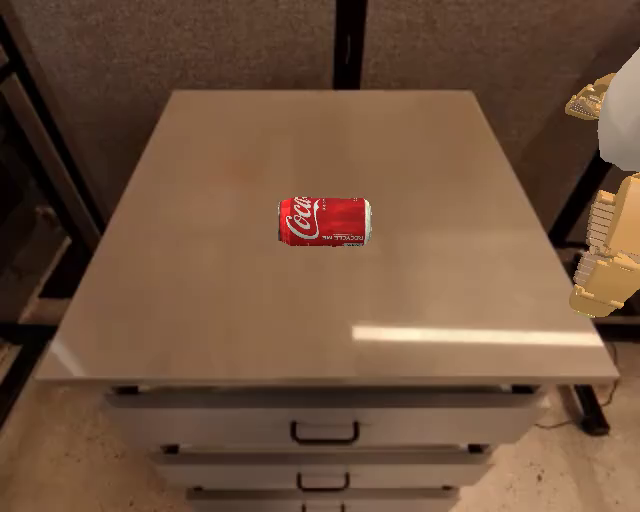}
        \par\smallskip
        {\footnotesize (a)\par}
    \end{minipage}\hfill
    \begin{minipage}[t]{0.49\textwidth}
        \centering
        \includegraphics[width=\linewidth]{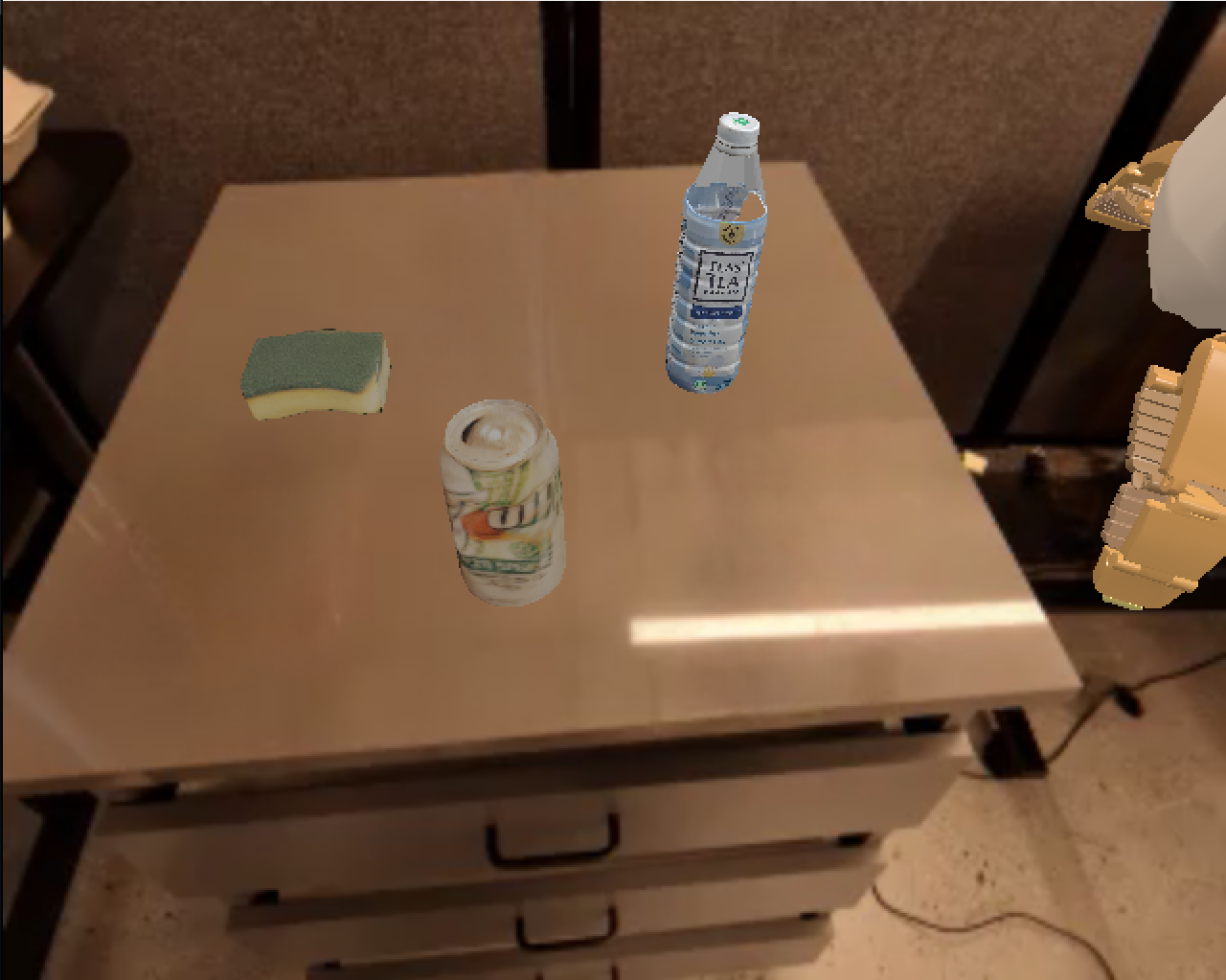}
        \par\smallskip
        {\footnotesize (b)\par}
    \end{minipage}
    \caption{Example visual observations from SimplerEnv:
    (a) a can on a tabletop for a \textit{``Pick-up''} task; and
    (b) a sponge, a can, and a bottle on a tabletop for a \textit{``Move Near''} task.}
    \label{fig:simplerenv-scenes}
\end{figure*}

Following prior work~\cite{valle2025evaluating, valle2026metamorphic, kanwal2026fatevla,wang2025vlatest}, we conducted our experiments using SimplerEnv~\cite{li24simpler}, an open-source benchmark for evaluating robot manipulation policies in simulation. Built on SAPIEN and ManiSkill2, SimplerEnv reproduces real-world robotic manipulation setups, providing a controlled environment in which experiments can be repeated under consistent conditions. This facilitates systematic execution of test cases and the comparison of the SBFL techniques across the evaluated VLA models. Figure~\ref{fig:simplerenv-scenes} shows example visual observations from the simulated manipulation environments. Within this benchmark, we used the Google Robot embodiment~\footnote{https://x.company/projects/everyday-robots/}, also known as the Everyday Robot, together with VLA checkpoints trained on the Fractal dataset~\cite{brohan2022rt}. At each decision step, the VLA policy processes visual observations, the natural-language task instruction, and the robot's proprioceptive state to generate actions. The simulated robot executes these actions, producing changes in both its position and the surrounding environment that are reflected in subsequent observations. This interaction continues until the task is completed or execution reaches its predefined limit, i.e., 80 steps for the ``\textit{Pick-up}'' task and 120 steps for the ``\textit{Move-Near}'' task. Using the same embodiment across experiments provides a common robotic platform for assessing our approach, while allowing different VLA policies to control the robot.

We organized the task selection around the objectives of the RQs. For RQ1--RQ4, we used the ``\textit{Pick-up}'' task for both training and testing the SBFL techniques. This task requires the robot to grasp the target object and lift it by at least $0.02$~m for five consecutive simulation frames. These experiments assess the techniques within a common task setting. Building on this evaluation, RQ5 investigates whether their effectiveness transfers to a different task. Both tasks used the same robotic embodiment, consistent with the Fractal dataset on which the VLA checkpoints were trained. This setup allows us to assess the cross-task generalizability of our approach while keeping the embodiment unchanged.

\subsection{Selected Baselines} 

We compared VoxelSBFL against eight trajectory-level uncertainty and quality metrics proposed by Valle et al.~\cite{valle2025evaluating}. We only used black-box techniques, as VoxelSBFL is also black-box. On the other hand, we opted not to include white-box uncertainty techniques from~\cite{valle2025evaluating} because (1) they included some overhead, which prevented us from executing a large set of trajectories and performing a comprehensive evaluation, and (2) each VLA would require the adaptation of such metrics, which is non-trivial. The uncertainty baselines comprised three action-level metrics: Action Position Instability, Action Velocity Instability, and Action Acceleration Instability. The quality baselines comprised TCP Position, Velocity, and Acceleration Instability, Trajectory Instability, and Optimal Trajectory. Because Optimal Trajectory requires information about the object and goal, we considered it a task-aware baseline. Valle et al.~\cite{valle2025evaluating} also proposed four token-level uncertainty metrics: Token Probability, Prediction Confidence Score (PCS), DeepGini, and Entropy. As mentioned above, we excluded these metrics because they require access to token-level prediction probabilities, which may depend on model-specific internal components and may not be available for every action-generation mechanism. Instead, we treat the VLA as a black box and consider information available at the interface between the policy and the robotic system, such as generated actions and executed trajectories. Therefore, for the rest of the paper, we call these black-box baseline metrics.

For each baseline, per-step and per-dimension values were reduced to a single trajectory-level score using their arithmetic mean. Score directions were fixed before evaluation, with larger instability, deviation, and distance values treated as stronger evidence of failure. A baseline was evaluated for a run only when its metric was available for every trajectory, thereby avoiding outcome-dependent episode removal. The primary cross-VLA comparison included only metrics available across every evaluated experimental cell.


\subsection{Generation of traces and spectra}

To evaluate VoxelSBFL, we sought successful and failed executions that varied in which parts of the workspace they visited. We adapted DeepHyperion~\cite{zohdinasab2021deephyperion} because its MAP-Elites search lets us pursue these two goals together. The search explores different initial object positions, which can elicit different reaching and grasping trajectories, and uses a failure-aware fitness function to prioritize scenes that expose failures within each region of the search grid. For each evaluated scene, we record the robot's TCP trajectory and the binary task outcome (i.e., pass or fail).
Each scene is represented as $s=(o,x,y,r)$, where $o$ denotes the manipulated object, $(x,y)$ its initial position, and $r$ its orientation. We consider seven objects and five predefined orientations, with $x\in[-0.5,-0.05]$ and $y\in[0,0.4]$ meters. MAP-Elites partitions the initial object positions into a $25\times25$ grid. Starting from 20 random scenes, the search mutates object positions and orientations. The fitness function prioritizes failures while also accounting for proximity to the object and spatial coverage.

We used four search configurations to examine whether the comparison of SBFL techniques depended on how the traces were generated. The baseline (\texttt{DH}) uses the MAP-Elites archive to guide exploration. The other configurations share the same scene representation and fitness function but guide candidate selection using a decision tree (\texttt{DH\_HybridDT}), a random forest (\texttt{DH\_HybridRF}), or an adaptive weighting of random-forest failure predictions, diversity, and novelty (\texttt{DH\_WeightedRF}). The predictors are trained on previously evaluated scenes and their observed outcomes. Within each experimental configuration, all SBFL techniques are evaluated on the same trajectories. For RQ1, we also recompute the aggregate ranking after excluding each generator in turn.

\subsection{Evaluation Metrics}

The primary evaluation metric was the area under the receiver operating characteristic curve (ROC-AUC), considering failed trajectories as the positive class. ROC-AUC measures whether the trajectory scores derived from the voxel suspiciousness map rank failed executions above successful ones. It is threshold-independent and is therefore suitable for comparing techniques, VLA models, and voxel sizes for which selecting a universal suspiciousness threshold would be arbitrary and potentially unfair. Conventional SBFL studies frequently use the EXAM score, defined as the percentage of ranked program elements that must be inspected before reaching the first known faulty element~\cite{wong2016survey}. EXAM assumes the existence of a ground-truth faulty element. This assumption does not hold in our dataset: a voxel is a spatial region rather than an intrinsically faulty executable element, failures may result from interactions across multiple regions and time steps, and no causal failure voxel is annotated. Consequently, the rank of the ``first faulty voxel'' is undefined. Our ROC-AUC results therefore evaluate failure-associated trajectory discrimination, instead of causal voxel localization. 




For RQ3, we also examine changes in technique rankings across voxel sizes, pairwise rank correlations, and the computational and sparsity costs associated with each resolution. For RQ4, we report the achieved calibration and test coverage and the matched difference in ROC-AUC between high- and low-coverage training subsets.

\subsection{Statistical tests}

Statistical comparisons were based on matched experimental observations. ROC-AUC values were first averaged across the ten leave-one-run-out folds within each configuration and then equally averaged across the two scene budgets. This yielded 16 matched blocks, corresponding to the combinations of four VLA models and four trajectory-generation strategies, which were used for the primary paired comparisons.
When more than two conditions were compared, we used the Friedman test as an omnibus test, followed, where appropriate, by paired two-sided exact Wilcoxon signed-rank tests. We applied Holm correction to account for multiple comparisons within separately defined families of hypotheses. For RQ1, the correction covered all 561 pairwise comparisons among the 34 SBFL techniques. The remaining RQs used separate correction families corresponding to their respective statistical comparisons. Alongside the Holm-adjusted $p$-values, we reported paired rank-biserial correlations ($r_{rb}$) as effect sizes. For RQ1, RQ2 and RQ3, we estimated pointwise 95\% confidence intervals using a hierarchical percentile bootstrap with 10,000 repetitions, resampling experimental blocks and validation folds while preserving paired comparisons. We considered absolute ROC-AUC differences below $0.01$ practically negligible when they were not statistically significant after Holm correction. This criterion was used to describe practical similarity rather than establish formal statistical equivalence.



\subsection{Dataset Construction}

The dataset contains episode metadata and time-ordered robot trajectories. Each episode includes a globally unique identifier, the VLA model, trajectory-generation condition, scene budget, run identifier, manipulation target, number of recorded steps, and binary pass/fail outcome. Each trajectory consists of a sequence of tool-center-point positions $(x,y,z)$ represented in the robot-base coordinate frame.
A fixed grid origin of $(-1,-1,0)$ was used.  
Unless voxel size was the factor under study, we used the reference size $s=0.02$ coordinate units.

\subsubsection{Dataset for RQ1 and RQ3:}
The dataset used for RQ1 and RQ3 contains 96,000 trajectories and 7,680,000 recorded Cartesian positions. It comprises four VLA models, four trajectory-generation conditions, two scene budgets (200 and 400 scenes), and ten independent runs for every model--condition--budget combination. This results in 32 configurations and 320 leave-one-run-out folds.

RQ1 used the reference voxel size of $0.02$. For every configuration, the suspiciousness map is trained using nine complete runs and evaluated on the remaining run. RQ3 reused the same trajectories, outcomes, and held-out-run identities at voxel sizes $0.01$, $0.02$, and $0.05$. The resulting matched design contains 960 voxel-size--fold cells and 32,640 technique--size--fold measurements.

\subsubsection{Dataset for RQ4:} RQ4 used the same trajectory collection but varied the number of trajectories used to construct the spectra. For each outer fold, one complete run was reserved as the test set, the next run in cyclic order was used as a calibration set, and the remaining eight runs formed the candidate training pool. The test run was never used for training or subset construction. The calibration set was used only to quantify the voxel coverage achieved by each training subset; it did not influence subset selection.

We considered training-set sizes of 10, 20, 50, 100, and 200 trajectories. For every configuration and outer fold, we generated ten repeated sequences of nested subsets. Within each repetition, successful and failed trajectories were randomly ordered separately, and a subset of size \(n\) contained the first \(n/2\) trajectories from each class. Consequently, every subset contained exactly 50\% failed and 50\% successful trajectories, and each smaller subset was contained in the subsequent larger subset. Calibration coverage was measured as

\[
    \operatorname{Coverage}(V_{\mathrm{train}})
    =
    \frac{
        \left|V_{\mathrm{train}}
        \cap
        V_{\mathrm{calibration}}\right|
    }{
        \left|V_{\mathrm{calibration}}\right|
    }.
\]

All 34 SBFL techniques were fitted to each subset and evaluated on the untouched test run. The resulting design comprised 32 VLA--trajectory-generation--budget configurations, ten outer folds, ten sampling repetitions, and five training-set sizes. This produced 16,000 fitted experimental cells and 544,000 technique-level measurements. This design treated training-set size as the controlled experimental factor and voxel coverage as an observed consequence of increasing the number of available trajectories.

\subsubsection{Dataset for RQ5:}

RQ5 evaluated the transfer of suspiciousness maps from the Pick-Up task to the Move-Near task. We used the maps reconstructed from the available Pick-Up training data, without retraining or calibrating them on Move-Near. For each VLA model, all 34 SBFL techniques were evaluated on the same selected Move-Near executions. No Move-Near execution was used to construct the suspiciousness maps.

The Move-Near working space comprised 72 voxels. Although our initial target was approximately 500 task instances per VLA model, we generated and executed 504 instances to ensure equal coverage of the working space: seven scenarios per voxel. This design provided balanced spatial coverage and increased the diversity of the evaluated scenarios. Task instances were generated using fixed random seeds, and, the same scene specifications were evaluated across VLA models. For each execution, we recorded the TCP trajectory and binary pass/fail outcome.

We then selected evaluation subsets from these executions using the proportions of successful and failed Pick-Up executions in RQ1 as a reference. For each model, these proportions were calculated by pooling the execution records from the 200- and 400-execution budgets.

Let \(q_m^{\mathrm{RQ1}}\) denote the smaller of the pass and fail proportions for model \(m\). We calculated the reference sample size corresponding to 100 expected observations of the minority class:


\[
    N_m^{\min}
    =
    \left\lceil \frac{100}{q_m^{\mathrm{RQ1}}} \right\rceil.
\]

For each VLA model, we then selected exactly 100 successful executions and the number of failed executions required to approximate the corresponding RQ1 failure proportion. Selection used a fixed random seed, with spatial constraints to maintain coverage across the 72 cells. The resulting evaluation subsets contained 298 executions for Pi0, 246 for GR00T, 223 for EO1, and 342 for SpatialVLA. Table~\ref{tab:target-selected-sample} reports the reference proportions, calculated sample sizes, and selected evaluation subsets.

\begin{table}[ht]
    \centering
    \caption{Pooled RQ1 success proportions and selected target-task evaluation samples.
    \(N_m^{\min}\) denotes the reference budget required for at least 100 expected
    observations of the minority class under the pooled RQ1 proportions.}
    \label{tab:target-selected-sample}
    \begin{tabular}{lrrrrr}
        \hline
        VLA & RQ1 pass (\%) & \(N_m^{\min}\) & Selected pass &
        Selected fail & Selected \(N_m\) \\
        \hline
        \(\pi_0\)  & 33.592 & 298 & 100 & 198 & 298 \\
        GR00T      & 40.654 & 246 & 100 & 146 & 246 \\
        EO1        & 44.875 & 223 & 100 & 123 & 223 \\
        SpatialVLA & 29.246 & 342 & 100 & 242 & 342 \\
        \hline
    \end{tabular}
\end{table}




\section{Analysis of Results and Discussion}
\label{sec:ExpResult}
\subsection{RQ1 -- Overall SBFL Technique Performance}


Table~\ref{tab:rq1-cross-vla} presents the macro ROC-AUC obtained by the 34 SBFL techniques for each VLA model at the primary voxel size. Harmonic Mean and Rogot2 jointly achieved the highest performance overall, with a macro ROC-AUC of 0.704, and ranked first for every VLA model. Their performance, nevertheless, varied across VLAs: their ROC-AUC ranged from 0.665 for EO1 to 0.746 for Pi0. This variation indicated that the separability of successful and failed trajectories depended partly on the VLA model, even though the identity of the leading techniques remained stable. Dennis exhibited comparatively
consistent behavior, ranking among the five best techniques for all four VLAs, while Phi ranked among the top five for GR00T, Pi0, and SpatialVLA but ranked sixth for EO1. In contrast, some techniques were more model-dependent. Tarantula ranked third for EO1 but sixth for GR00T, ninth for Pi0, and eighth for SpatialVLA. Similarly, Tarwid ranked fourth for EO1 and fifth for GR00T, but only tenth for Pi0 and seventh for SpatialVLA. Zoltar ranged from fifth for SpatialVLA to fifteenth for EO1. These differences suggested that the distribution of passing and failing voxel spectra varied across VLA models and
affected the formulas differently.

The table also revealed substantial differences among the 34 techniques. The techniques occupying the top five overall ranks obtained macro ROC-AUC values between 0.655 and 0.704, whereas several techniques performed close to random discrimination, with some obtaining ROC-AUC values below 0.5 for at least one VLA. Tarantula and Tarwid also produced nearly identical overall results, both obtaining a macro ROC-AUC of 0.655 at the reported precision. Moreover, several
groups of techniques produced identical or nearly identical results, including Hamann, Hamming, Rogers \& Tanimoto, Simple Matching, and Sokal, as well as Anderberg, Dice, Goodman, Jaccard, Kulczynski, and Sorensen-Dice. This redundancy may have resulted from mathematical relationships among the formulas or from different formulas inducing the same voxel ordering under sparse and highly tied spectra. Similar redundancies among SBFL techniques have also been reported in previous comparative studies~\cite{arrieta2018spectrum}.

\begin{table}[t]
\centering
\caption{RQ1 -- Macro ROC-AUC of the 34 SBFL techniques for each VLA model. Parentheses report tie-aware ranks. Bold denotes rank 1 and underlining denotes the remaining top-five ranks. Harmonic Mean and Rogot2 are tied because they produce identical results in every reported column.}
\label{tab:rq1-cross-vla}
{\scriptsize
\setlength{\tabcolsep}{4pt}
\renewcommand{\arraystretch}{0.96}
\begin{tabular}{lccccc}
\toprule
Technique & EO1 & GR00T & Pi0 & SpatialVLA & Overall \\
\midrule
Harmonic Mean & \textbf{0.665 (1)} & \textbf{0.673 (1)} & \textbf{0.746 (1)} & \textbf{0.732 (1)} & \textbf{0.704 (1)} \\
Rogot2 & \textbf{0.665 (1)} & \textbf{0.673 (1)} & \textbf{0.746 (1)} & \textbf{0.732 (1)} & \textbf{0.704 (1)} \\
Dennis & \underline{0.646 (5)} & \underline{0.660 (3)} & \underline{0.738 (3)} & \underline{0.708 (3)} & \underline{0.688 (3)} \\
Phi (Geometric Mean) & 0.638 (6) & \underline{0.653 (4)} & \underline{0.737 (4)} & \underline{0.699 (4)} & \underline{0.682 (4)} \\
Tarantula & \underline{0.659 (3)} & 0.645 (6) & 0.671 (9) & 0.644 (8) & \underline{0.655 (5)} \\
Tarwid & \underline{0.659 (4)} & \underline{0.645 (5)} & 0.671 (10) & 0.644 (7) & 0.655 (6) \\
Arithmetic Mean & 0.628 (7) & 0.622 (7) & \underline{0.711 (5)} & 0.651 (6) & 0.653 (7) \\
Michael & 0.625 (8) & 0.618 (8) & 0.707 (6) & 0.631 (9) & 0.645 (8) \\
Cohen & 0.619 (9) & 0.618 (9) & 0.707 (7) & 0.628 (10) & 0.643 (9) \\
Zoltar & 0.579 (15) & 0.599 (10) & 0.703 (8) & \underline{0.690 (5)} & 0.643 (10) \\
Ochiai2 & 0.605 (10) & 0.563 (12) & 0.667 (11) & 0.521 (12) & 0.589 (11) \\
Mountford & 0.556 (21) & 0.574 (11) & 0.648 (12) & 0.566 (11) & 0.586 (12) \\
Rogot1 & 0.597 (11) & 0.521 (21) & 0.610 (13) & 0.495 (18) & 0.556 (13) \\
Scott & 0.597 (11) & 0.521 (21) & 0.610 (13) & 0.495 (18) & 0.556 (13) \\
Fleiss & 0.595 (13) & 0.521 (23) & 0.593 (15) & 0.495 (20) & 0.551 (15) \\
Ample & 0.581 (14) & 0.522 (19) & 0.533 (18) & 0.481 (32) & 0.529 (16) \\
Hamann & 0.557 (16) & 0.529 (13) & 0.474 (19) & 0.496 (13) & 0.514 (17) \\
Hamming & 0.557 (16) & 0.529 (13) & 0.474 (19) & 0.496 (13) & 0.514 (17) \\
Rogers \& Tanimoto & 0.557 (16) & 0.529 (13) & 0.474 (19) & 0.496 (13) & 0.514 (17) \\
Simple Matching & 0.557 (16) & 0.529 (13) & 0.474 (19) & 0.496 (13) & 0.514 (17) \\
Sokal & 0.557 (16) & 0.529 (13) & 0.474 (19) & 0.496 (13) & 0.514 (17) \\
Pearson & 0.537 (22) & 0.521 (20) & 0.568 (16) & 0.399 (33) & 0.506 (22) \\
Baroni-Urbani \& Buser & 0.525 (23) & 0.486 (24) & 0.414 (26) & 0.494 (23) & 0.480 (23) \\
Gower & 0.455 (34) & 0.526 (18) & 0.548 (17) & 0.384 (34) & 0.478 (24) \\
Braun-Banquet & 0.517 (24) & 0.472 (27) & 0.414 (33) & 0.494 (30) & 0.474 (25) \\
Ochiai & 0.504 (25) & 0.472 (25) & 0.414 (24) & 0.495 (21) & 0.471 (26) \\
Fossum & 0.504 (25) & 0.472 (26) & 0.414 (24) & 0.495 (21) & 0.471 (27) \\
Anderberg & 0.504 (27) & 0.472 (28) & 0.414 (27) & 0.494 (24) & 0.471 (28) \\
Dice & 0.504 (27) & 0.472 (28) & 0.414 (27) & 0.494 (24) & 0.471 (28) \\
Goodman & 0.504 (27) & 0.472 (28) & 0.414 (27) & 0.494 (24) & 0.471 (28) \\
Jaccard & 0.504 (27) & 0.472 (28) & 0.414 (27) & 0.494 (24) & 0.471 (28) \\
Kulczynski & 0.504 (27) & 0.472 (28) & 0.414 (27) & 0.494 (24) & 0.471 (28) \\
Sorensen-Dice & 0.504 (27) & 0.472 (28) & 0.414 (27) & 0.494 (24) & 0.471 (28) \\
Pierce & 0.504 (33) & 0.470 (34) & 0.413 (34) & 0.494 (31) & 0.470 (34) \\
\bottomrule
\end{tabular}
}
\end{table}
 
Table~\ref{tab:rq1-competitive-statistics} presents the statistical comparison of the techniques that reached a top-five rank for at least one VLA. Rogot2 reproduced exactly the performance of Harmonic Mean: its ROC-AUC gap was zero across all 16 matched experimental blocks, its Holm-adjusted $p$-value was 1.000, and its paired rank-biserial effect size was zero. Consequently, both techniques achieved identical failure-discrimination performance under the evaluated conditions. Dennis and Phi were the closest alternatives, with ROC-AUC gaps of 0.016 and 0.022, respectively. However, neither difference remained
statistically significant after Holm correction ($p=0.279$ and $p=0.134$, respectively). In contrast, Tarantula, Tarwid, Arithmetic Mean, and Zoltar performed significantly worse than Harmonic Mean, with adjusted $p$-values between 0.017 and 0.018. The 95\% confidence intervals for Dennis and Phi excluded zero, although these pointwise intervals were not adjusted for multiple comparisons. Therefore, their observed performance differences should not be interpreted as statistically significant under the corrected tests.

\begin{table}[t]
\centering
\caption{RQ1 -- Statistical comparison of the techniques reaching a top-five rank for at least one VLA. The AUC gap is the leader's ROC-AUC minus the candidate's ROC-AUC. Confidence intervals are pointwise 95\%; $p$-values are adjusted using Holm correction over all 561 pairwise comparisons among the 34 SBFL techniques, and $r_{rb}$ is the paired rank-biserial effect size.}
\label{tab:rq1-competitive-statistics}
{\scriptsize
\setlength{\tabcolsep}{4pt}
\begin{tabular}{lccccc}
\toprule
Technique & ROC-AUC [95\% CI] & Gap [95\% CI] & Holm $p$ & $r_{rb}$ & Equivalent \\
\midrule
\textbf{Harmonic Mean}
    & 0.704 [0.680, 0.728]
    & Ref. & -- & -- & Ref. \\
\textbf{Rogot2}
    & 0.704 [0.680, 0.728]
    & 0.000 [0.000, 0.000]
    & 1.000 & 0.000 & Yes \\
Dennis
    & 0.688 [0.667, 0.710]
    & 0.016 [0.007, 0.025]
    & 0.279 & 0.838 & No \\
Phi (Geometric Mean)
    & 0.682 [0.659, 0.705]
    & 0.022 [0.012, 0.032]
    & 0.134 & 0.882 & No \\
Tarantula
    & 0.655 [0.636, 0.675]
    & 0.049 [0.031, 0.069]
    & 0.018 & 0.985 & No \\
Tarwid
    & 0.655 [0.636, 0.675]
    & 0.049 [0.031, 0.069]
    & 0.018 & 0.985 & No \\
Arithmetic Mean
    & 0.653 [0.632, 0.676]
    & 0.051 [0.036, 0.067]
    & 0.017 & 1.000 & No \\
Zoltar
    & 0.643 [0.614, 0.672]
    & 0.061 [0.049, 0.075]
    & 0.017 & 1.000 & No \\
\bottomrule
\end{tabular}
}
\end{table}

Overall, Harmonic Mean and Rogot2 consistently achieved the highest observed ROC-AUC across all four VLA models. Their joint leading position was also preserved when each VLA model or trajectory-generation strategy was excluded in turn, indicating that the overall ranking was not driven by a single experimental group. Across the complete family of pairwise comparisons, Harmonic Mean showed statistically significant differences against 30 of the remaining 33 techniques, whereas the comparisons against Rogot2, Dennis, and Phi were not significant after Holm correction. These results supported retaining Harmonic Mean as the reference technique for the subsequent research questions, while acknowledging Rogot2 as an equally effective choice under the primary experimental conditions. Dennis and Phi remained the strongest alternatives in terms of observed ROC-AUC, although the corrected statistical analysis did not establish a significant difference from the leading techniques.
Overall, the first RQ can be answered as follows: 

\begin{custombox}{Answer to RQ1}
Under the studied conditions, Harmonic Mean and Rogot2 achieved the highest observed failure-discrimination performance among the 34 evaluated SBFL techniques, jointly ranking first for every VLA model with an overall macro ROC-AUC of 0.704. The corrected statistical analysis showed significant differences between Harmonic Mean and 30 of the other 33 techniques, but not Rogot2, Dennis, or Phi.
\end{custombox}

\subsection{RQ2 -- Baseline Comparison}


Table~\ref{tab:rq2-baseline-performance} compares Harmonic Mean, the VoxelSBFL reference technique selected in RQ1, with the black-box metrics proposed by Valle et al.~\cite{valle2025evaluating}. These metrics capture different aspects of VLA behavior, including changes in generated actions, instability in the executed trajectory, and progress towards task-relevant positions. The comparison allows us to assess whether the spatial information captured by VoxelSBFL provides effective discrimination between successful and failed executions. 
Paired Wilcoxon signed-rank tests with Holm correction showed that Harmonic Mean significantly outperformed all seven action- and motion-instability baselines ($p_{\mathrm{Holm}} \leq 0.011$). Optimal Trajectory, however, significantly outperformed Harmonic Mean ($p_{\mathrm{Holm}} = 0.0023$).

\begin{table*}[ht]
    \centering
    \caption{RQ2 -- Trajectory-level failure-discrimination ROC-AUC. Harmonic Mean is the VoxelSBFL reference technique selected in RQ1. Bold denotes the best result and underlining denotes the second-best result in each column. \(^{\dagger}\) denotes a task-aware metric using object and goal information and $^{\star}$ indicates an SBFL technique .}
    \label{tab:rq2-baseline-performance}
    \resizebox{0.75\textwidth}{!}{
    \begin{tabular}{lccccc}
        \toprule
        Method & EO1 & GR00T & Pi0 & SpatialVLA & Overall \\
        \midrule
        Optimal Trajectory$^{\dagger}$ & 0.6666 & \textbf{0.7131} & \textbf{0.8029} & \textbf{0.8408} & \textbf{0.7559} \\
        Harmonic Mean$^{\star}$ & 0.6652 & \underline{0.6730} & \underline{0.7461} & \underline{0.7315} & \underline{0.7039}  \\
        Rogot2$^{\star}$ & 0.6652 & \underline{0.6730} & \underline{0.7461} & \underline{0.7315} & \underline{0.7039}  \\
        Dennis$^{\star}$ & 0.6462 & 0.6602 & 0.7383 & 0.7075 & 0.6880  \\
        Phi (Geometric Mean)$^{\star}$ & 0.6375 & 0.6533 & 0.7370 & 0.6994 & 0.6818  \\
        TCP Acceleration Instability & \textbf{0.6995} & 0.5598 & 0.6671 & 0.6934 & 0.6550  \\
        Tarantula$^{\star}$ & 0.6590 & 0.6451 & 0.6715 & 0.6439 & 0.6549  \\
        TCP Velocity Instability & 0.6782 & 0.5964 & 0.6136 & 0.6745 & 0.6407 \\
        Trajectory Instability & \underline{0.6891} & 0.5432 & 0.6195 & 0.6750 & 0.6317  \\
        TCP Position Instability & 0.6124 & 0.6496 & 0.5624 & 0.5965 & 0.6052  \\
        Action Position Instability & 0.5518 & 0.5796 & 0.5571 & 0.7038 & 0.5981  \\
        Action Velocity Instability & 0.5614 & 0.5599 & 0.5606 & 0.6992 & 0.5953  \\
        Action Acceleration Instability & 0.5569 & 0.5555 & 0.5569 & 0.6941 & 0.5908  \\
        \bottomrule
    \end{tabular}
    }
\end{table*}

At the aggregate level, Optimal Trajectory achieved the highest macro ROC-AUC, followed jointly by Harmonic Mean and Rogot2, which produced identical reported scores for every VLA model. Dennis and Phi (Geometric Mean) followed in fourth and fifth overall ranking, respectively. All these four SBFL techniques outperformed every reported action- and motion-instability baseline overall. Among these baselines, TCP Acceleration Instability achieved the highest value, followed by TCP Velocity Instability and Trajectory Instability. However, this advantage did not extend to every SBFL technique. For instance, Tarantula achieved nearly matching scores to TCP Acceleration Instability at the reported precision. These results suggest that information about \emph{where} the robot moves (VoxelSBFL) can complement information about \emph{how} (our previous work \cite{valle2025evaluating}) it moves when distinguishing successful and failed executions. The instability metrics proposed in our previous study \cite{valle2025evaluating} characterize variations in actions or motion, whereas VoxelSBFL uses previously labeled executions to associate workspace regions with failures. Consequently, a trajectory may receive a high suspiciousness score based on VoxelSBFL because it visits regions associated with failures. However, its movements can be soft and not exhibit pronounced instability. 

Relative performance nevertheless varied across VLA models. For GR00T, Pi0, and SpatialVLA, Optimal Trajectory achieved the highest ROC-AUC, while Harmonic Mean and Rogot2 shared second place. Both techniques outperformed every action- and motion-instability metric baseline for these three models. Dennis also exceeded all instability baselines for these models. Phi similarly exceeded all instability baselines for GR00T and Pi0, but its SpatialVLA result was below Action Position Instability. Tarantula showed a less consistent advantage, exceeding all instability baselines for Pi0 but not for the other models. Nevertheless, EO1 exhibited a different pattern. TCP Acceleration Instability achieved the highest ROC-AUC score, followed by Trajectory Instability and TCP Velocity Instability. Thus, the three strongest motion-instability metrics outperformed both the Optimal Trajectory metric and all reported SBFL techniques for EO1. This pattern suggests that motion characteristics were more informative for failure discrimination in this model under the evaluated conditions. A plausible explanation might be that passing and failing trajectories visited similar regions but differed in their movements within those regions.

Optimal Trajectory provided the strongest overall discrimination. However, its application requires task-specific information to estimate the optimal trajectory, including the position of the target object and, where applicable, the goal. Consequently, its application in real-world operation depends on the availability of this information, which may need to be estimated rather than obtained directly from the environment. In contrast, VoxelSBFL does not require explicit object or goal positions, but only suspiciousness maps constructed from previously labeled executions. The comparison therefore reflects differences in both predictive performance and the information required by each method.



\begin{custombox}{Answer to RQ2}
Harmonic Mean and Rogot2 achieved higher overall ROC-AUC than all evaluated action- and motion-instability baselines. The advantage of Harmonic Mean and Rogot2 was statistically significant after Holm correction Only Optimal Trajectory achieved significantly higher performance, but it requires task-specific object and goal information that may not be available in real-world operation. The leading SBFL techniques therefore offer a competitive alternative to the evaluated black-box metrics, although their relative performance varies across models.
\end{custombox}

\subsection{RQ3 -- Voxel-Size Sensitivity:}

Table~\ref{tab:RQ3-reference-by-vla} reports the sensitivity of Harmonic Mean, the reference technique selected in RQ1, to the voxel size. At the aggregate level, the primary size of 0.02 achieved the highest macro ROC-AUC (0.7039), followed by 0.01 (0.6946) and 0.05 (0.6930). However, the descriptively best size varied across VLA models. EO1 and SpatialVLA performed best with 0.01, Pi0 performed best with 0.02, and GR00T performed best with 0.05. The magnitude and direction of the changes also differed: increasing the voxel size reduced the performance of EO1 and SpatialVLA, whereas it progressively improved the performance of GR00T. These results indicate that no voxel size was descriptively optimal for every VLA. This model-specific behavior could have resulted from differences in the spatial distributions explored by their trajectories: models producing more concentrated trajectories might have benefited from finer spatial distinctions, whereas models covering more variable regions might have benefited from the aggregation provided by larger voxels.

\begin{table}[ht]
\centering
\caption{RQ3 -- Voxel-size sensitivity of Harmonic Mean, the frozen RQ1 reference technique. Values are macro ROC-AUC over the four trajectory-generation conditions for each VLA. Bold denotes the descriptive best size within a row; inferential tests are reported separately in Table~\ref{tab:RQ3-global-statistics}.}
\label{tab:RQ3-reference-by-vla}
\begin{tabular}{lcccc}
\toprule
VLA & 0.01 & 0.02 & 0.05 & Descriptive best size \\
\midrule
EO1 & \textbf{0.6711} & 0.6652 & 0.6253 & 0.01 \\
GR00T & 0.6574 & 0.6730 & \textbf{0.6815} & 0.05 \\
Pi0 & 0.7101 & \textbf{0.7461} & 0.7416 & 0.02 \\
SpatialVLA & \textbf{0.7398} & 0.7315 & 0.7235 & 0.01 \\
\midrule
\textbf{Macro} & 0.6946 & \textbf{0.7039} & 0.6930 & 0.02 \\
\bottomrule
\end{tabular}
\end{table}


Table~\ref{tab:RQ3-techniques-by-vla} shows how voxel size affected the relative performance of the techniques that reached a top-five rank for at least one size within each VLA. At the primary size of 0.02, Harmonic Mean and Rogot2 jointly ranked first for all four VLAs. This consistency was not always preserved when voxel size varied. At 0.01, Harmonic Mean and Rogot2 remained the leaders for EO1, but Dennis became the best technique for GR00T and Pi0, while Zoltar ranked first for SpatialVLA. The most pronounced change occurred at 0.05, where Tarantula ranked first for EO1, GR00T, and Pi0, whereas Tarwid ranked first for SpatialVLA. The two techniques produced nearly identical results at the reported precision, with ROC-AUC values between 0.6877 and 0.7649, and their different ranks resulted from small differences in the unrounded values. However, both techniques performed poorly at 0.01: Tarantula ranked between tenth and sixteenth, while Tarwid ranked between tenth and fifteenth. Zoltar exhibited the opposite pattern, ranking among the five best techniques for every VLA at 0.01 but falling between fourteenth and thirty-second at 0.05. Therefore, voxel size did not merely shift absolute ROC-AUC values; it also changed the relative suitability of the suspiciousness formulas.
A plausible explanation is that coarse voxels combined visits from more trajectories and produced more stable class-conditional evidence, which may have favored techniques such as Tarantula and Tarwid. In contrast, finer partitions preserved localized distinctions that may have benefited Zoltar.

\begin{table*}[t]
\centering
\caption{RQ3 -- Voxel-size sensitivity of techniques reaching a
top-five rank for at least one size within each VLA. Each cell reports
macro ROC-AUC and the rank in parentheses. Ranks use unrounded values.
Bold denotes rank 1 and underlining denotes the remaining top-five
ranks. }
\label{tab:RQ3-techniques-by-vla}
{\scriptsize
\setlength{\tabcolsep}{7pt}
\renewcommand{\arraystretch}{0.96}
\begin{tabular}{lccc}
\toprule
Technique & 0.01 ROC-AUC (rank) & 0.02 ROC-AUC (rank) & 0.05 ROC-AUC (rank) \\
\midrule
\multicolumn{4}{l}{\textit{(a) EO1}} \\
Harmonic Mean & \textbf{0.6711 (1)} & \textbf{0.6652 (1)} & \underline{0.6253 (5)} \\
Rogot2 & \textbf{0.6711 (1)} & \textbf{0.6652 (1)} & \underline{0.6253 (4)} \\
Tarantula & 0.5460 (16) & \underline{0.6590 (3)} & \textbf{0.7089 (1)} \\
Tarwid & 0.5460 (15) & \underline{0.6590 (4)} & \underline{0.7089 (2)} \\
Dennis & \underline{0.6555 (4)} & \underline{0.6462 (5)} & \underline{0.6680 (3)} \\
Zoltar & \underline{0.6705 (3)} & 0.5794 (15) & 0.4999 (32) \\
Phi (Geometric Mean) & \underline{0.5977 (5)} & 0.6375 (6) & 0.6204 (6) \\
\midrule
\multicolumn{4}{l}{\textit{(b) GR00T}} \\
Harmonic Mean & \underline{0.6574 (2)} & \textbf{0.6730 (1)} & \underline{0.6815 (4)} \\
Rogot2 & \underline{0.6574 (2)} & \textbf{0.6730 (1)} & \underline{0.6817 (3)} \\
Dennis & \textbf{0.6594 (1)} & \underline{0.6602 (3)} & \underline{0.6668 (5)} \\
Tarantula & 0.5289 (11) & 0.6451 (6) & \textbf{0.6877 (1)} \\
Tarwid & 0.5290 (10) & \underline{0.6451 (5)} & \underline{0.6877 (2)} \\
Phi (Geometric Mean) & \underline{0.6561 (4)} & \underline{0.6533 (4)} & 0.6627 (6) \\
Zoltar & \underline{0.6556 (5)} & 0.5985 (10) & 0.4999 (25) \\
\midrule
\multicolumn{4}{l}{\textit{(c) Pi0}} \\
Harmonic Mean & \underline{0.7101 (3)} & \textbf{0.7461 (1)} & \underline{0.7416 (3)} \\
Rogot2 & \underline{0.7101 (3)} & \textbf{0.7461 (1)} & \underline{0.7416 (3)} \\
Dennis & \textbf{0.7149 (1)} & \underline{0.7383 (3)} & \underline{0.7096 (5)} \\
Tarantula & 0.5152 (15) & 0.6715 (9) & \textbf{0.7649 (1)} \\
Phi (Geometric Mean) & \underline{0.7147 (2)} & \underline{0.7370 (4)} & 0.6705 (6) \\
Tarwid & 0.5152 (14) & 0.6714 (10) & \underline{0.7649 (2)} \\
Arithmetic Mean & 0.6989 (6) & \underline{0.7106 (5)} & 0.6517 (7) \\
Zoltar & \underline{0.7017 (5)} & 0.7031 (8) & 0.5402 (15) \\
\midrule
\multicolumn{4}{l}{\textit{(d) SpatialVLA}} \\
Harmonic Mean & \underline{0.7398 (2)} & \textbf{0.7315 (1)} & \underline{0.7235 (3)} \\
Rogot2 & \underline{0.7398 (2)} & \textbf{0.7315 (1)} & \underline{0.7235 (3)} \\
Zoltar & \textbf{0.7421 (1)} & \underline{0.6902 (5)} & 0.5452 (14) \\
Tarwid & 0.5113 (11) & 0.6439 (7) & \textbf{0.7564 (1)} \\
Tarantula & 0.5113 (10) & 0.6439 (8) & \underline{0.7564 (2)} \\
Dennis & \underline{0.7129 (4)} & \underline{0.7075 (3)} & \underline{0.7141 (5)} \\
Phi (Geometric Mean) & \underline{0.7115 (5)} & \underline{0.6994 (4)} & 0.6960 (6) \\
\bottomrule
\end{tabular}
}
\end{table*}


Table~\ref{tab:RQ3-diagnostics-by-vla} analyzes the coverage and computational consequences of changing the voxel size. The fraction of known voxels increased consistently with voxel size. At 0.01, known coverage ranged from 84.2\% for Pi0 to 92.8\% for GR00T; at 0.02, it ranged from 97.8\% to 98.8\%; and at 0.05, it reached between 99.7\% and 99.8\%. This improvement was accompanied by a substantial reduction in the number of training voxels. Depending on the VLA, the mean number decreased from approximately 31,000--38,000 at 0.01 to
5,700--7,500 at 0.02 and only 570--740 at 0.05. The complete 34-technique artifacts decreased from 1.58--2.86 MB at 0.01 to 0.15--0.18 MB at 0.05. Execution times decreased accordingly: fitting required 0.137--0.180 seconds at 0.01 and 0.025--0.034 seconds at 0.05, while scoring decreased from 0.244--0.296 seconds to 0.014--0.016 seconds. Thus, while larger voxels produce considerably more compact and computationally efficient models, smaller voxels remain fast for practical adoption.

\begin{table*}[t]
\centering
\caption{RQ3 -- Coverage and computational diagnostics by VLA and voxel size for Harmonic Mean. Values are averaged over the 80 held-out folds in each VLA-size cell. Artifact sizes use decimal megabytes.}
\label{tab:RQ3-diagnostics-by-vla}
{\scriptsize
\setlength{\tabcolsep}{4pt}
\begin{tabular}{lrrrrrrr}
\toprule
VLA & Size & Known & Ties & Train voxels & Artifact (MB) & Fit (s) & Score (s) \\
\midrule
EO1 & 0.01 & 92.4\% & 90.1\% & 30,791 & 2.12 & 0.146 & 0.244 \\
 & 0.02 & 98.8\% & 67.2\% & 5,651 & 0.93 & 0.061 & 0.057 \\
 & 0.05 & 99.8\% & 84.6\% & 572 & 0.15 & 0.028 & 0.014 \\
\midrule
GR00T & 0.01 & 92.8\% & 90.7\% & 37,545 & 2.86 & 0.180 & 0.296 \\
 & 0.02 & 98.8\% & 59.1\% & 7,231 & 1.18 & 0.080 & 0.071 \\
 & 0.05 & 99.8\% & 75.0\% & 736 & 0.18 & 0.034 & 0.016 \\
\midrule
Pi0 & 0.01 & 84.2\% & 93.3\% & 37,656 & 1.58 & 0.149 & 0.287 \\
 & 0.02 & 97.9\% & 69.2\% & 7,532 & 1.03 & 0.061 & 0.068 \\
 & 0.05 & 99.7\% & 71.9\% & 738 & 0.18 & 0.027 & 0.016 \\
\midrule
SpatialVLA & 0.01 & 86.9\% & 91.1\% & 33,879 & 1.69 & 0.137 & 0.264 \\
 & 0.02 & 97.8\% & 66.3\% & 6,773 & 0.88 & 0.057 & 0.064 \\
 & 0.05 & 99.7\% & 76.8\% & 692 & 0.16 & 0.025 & 0.015 \\
\bottomrule
\end{tabular}
}
\end{table*}


The score-tie results in Table~\ref{tab:RQ3-diagnostics-by-vla} exhibited a non-monotonic pattern. The finest voxel size produced the largest tie fractions, ranging from 90.1\% to 93.3\%. These fractions decreased substantially at 0.02 before increasing again at 0.05. This pattern suggested two
different sources of ambiguity. With 0.01 voxels, the large and sparsely covered voxel space may have caused many low-support voxels to share the same contingency counts and suspiciousness values, leading multiple trajectories to receive identical aggregated risk scores.
Conversely, at 0.05, aggregating many positions into a small number of voxels may have caused more trajectories to traverse voxels with similar spectra, again reducing the diversity of their risk scores. The intermediate size of 0.02 produced the lowest tie fraction for
every VLA while maintaining known coverage between 97.8\% and 98.8\%.
Descriptively, it therefore offered the most favorable compromise between spatial resolution, coverage, and trajectory-score discrimination for Harmonic Mean.
The exact paired comparisons identified ten significant technique-specific voxel-size effects after Holm correction, indicating that sensitivity to voxel size depended on the suspiciousness technique.

Table~\ref{tab:RQ3-global-statistics} showed that the differences observed for Harmonic Mean were not statistically significant at the global level. The Friedman test did not reject the null hypothesis of equal performance across voxel sizes ($\chi_F^2(2)=1.625$, $p=0.444$). Compared with 0.02, the ROC-AUC decreased by 0.0093 at 0.01 and by 0.0110 at 0.05, and both confidence intervals included zero. The 0.01 configuration was classified as practically equivalent to the reference size, whereas 0.05 fell outside the predefined practical-equivalence group. Therefore, the data did not provide sufficient evidence of a general voxel-size effect for Harmonic Mean, although 0.02 remained the descriptively strongest configuration.

\begin{table*}[t]
\centering
\caption{RQ3 -- Global statistical analysis over the 16 matched
VLA--trajectory-condition blocks. Panel (a) reports the frozen Harmonic
Mean size comparison (Friedman $\chi^2_F(2)=1.625$, $p=0.444$).
Panel (b) contains interactions satisfying Holm-corrected $p<0.05$ and
$|\Delta\Delta\mathrm{AUC}|\geq0.01$. Panel (c) reports tie-aware
Spearman rank correlations across all 34 techniques.}
\label{tab:RQ3-global-statistics}
{\scriptsize
\setlength{\tabcolsep}{4pt}
\begin{tabular}{lccc}
\toprule
\multicolumn{4}{l}{\textit{(a) Voxel-size effect for Harmonic Mean}} \\
Size & ROC-AUC [95\% CI] & $\Delta$ vs. 0.02 [95\% CI] & Equivalent \\
\midrule
0.01 & 0.6946 [0.6727, 0.7178] & -0.0093 [-0.0219, +0.0030] & Yes \\
0.02 & 0.7039 [0.6796, 0.7280] & +0.0000 [+0.0000, +0.0000] & Yes \\
0.05 & 0.6930 [0.6652, 0.7188] & -0.0110 [-0.0271, +0.0033] & No \\
\bottomrule
\end{tabular}

\vspace{2mm}
\begin{tabular}{llcccc}
\toprule
\multicolumn{6}{l}{\textit{(b) Practically meaningful technique--size interactions}} \\
Size vs. 0.02 & Technique & $\Delta$AUC &
$\Delta\Delta$AUC [95\% CI] & Holm $p$ & $r_{rb}$ \\
\midrule
0.01 & Mountford & +0.0506 & +0.0600 [+0.0342, +0.0895] & 0.002 & +1.000 \\
0.01 & Zoltar & +0.0496 & +0.0590 [+0.0467, +0.0721] & 0.002 & +1.000 \\
0.01 & Ochiai2 & -0.0658 & -0.0565 [-0.0690, -0.0438] & 0.004 & -0.985 \\
0.01 & Tarwid & -0.1295 & -0.1202 [-0.1338, -0.1068] & 0.002 & -1.000 \\
0.01 & Tarantula & -0.1295 & -0.1202 [-0.1338, -0.1068] & 0.002 & -1.000 \\
0.05 & Tarwid & +0.0746 & +0.0856 [+0.0615, +0.1123] & 0.002 & +1.000 \\
0.05 & Tarantula & +0.0746 & +0.0856 [+0.0615, +0.1123] & 0.002 & +1.000 \\
0.05 & Zoltar & -0.1215 & -0.1105 [-0.1340, -0.0848] & 0.004 & -0.985 \\
\bottomrule
\end{tabular}

\vspace{2mm}
\begin{tabular}{lc}
\toprule
\multicolumn{2}{l}{\textit{(c) Ranking stability}} \\
Voxel-size pair & Spearman $\rho$ \\
\midrule
0.01--0.02 & 0.921 \\
0.01--0.05 & 0.769 \\
0.02--0.05 & 0.899 \\
\bottomrule
\end{tabular}
}
\end{table*}

The interaction analysis nevertheless identified eight cases in which voxel size affected a technique differently from Harmonic Mean. At 0.01, Mountford and Zoltar improved relative to the primary size, with difference-in-differences values of 0.0600 and 0.0590, respectively. In contrast, Ochiai2 deteriorated at 0.01 ($\Delta\Delta\mathrm{AUC}=-0.0565$). Tarwid and Tarantula exhibited the strongest size-dependent responses: both deteriorated sharply at 0.01 ($\Delta\Delta\mathrm{AUC}=-0.1202$) but improved at 0.05
($\Delta\Delta\mathrm{AUC}=0.0856$). Zoltar exhibited the reverse response at 0.05 ($\Delta\Delta\mathrm{AUC}=-0.1105$). All eight interactions remained significant after Holm correction and showed very large paired effect sizes, with absolute $r_{rb}$ values between 0.985 and 1.000. The technique rankings were nevertheless relatively stable overall, particularly between 0.01 and 0.02 ($\rho=0.921$) and between 0.02 and 0.05 ($\rho=0.899$). The lower correlation between the two extreme sizes ($\rho=0.769$) indicated that ranking changes accumulated as the difference in spatial granularity increased.

Overall, voxel size did not produce a statistically significant global change in the performance of Harmonic Mean, although its descriptive effect varied across VLAs and several SBFL techniques responded differently to changes in granularity. The size of 0.02 achieved the highest aggregate ROC-AUC for the reference technique and offered high known coverage with substantially fewer voxels and fewer score ties than 0.01. Although 0.05 was the most computationally efficient configuration and strongly favored Tarantula and Tarwid, it reduced spatial resolution and was not practically equivalent to 0.02 for Harmonic Mean. Consequently, 0.02 descriptively represented the most balanced default for trajectory-level evaluation in the considered setting. However, the strong interactions observed for Tarantula, Tarwid, Zoltar, Mountford, and Ochiai2 demonstrated that voxel size needed to be controlled when comparing SBFL techniques. This selection
should not be interpreted as establishing 0.02 as the optimal localization resolution, which would require ground truth about the problematic spatial regions.

\begin{custombox}{Answer to RQ3}
No voxel size was uniformly optimal across all VLA models and SBFL techniques. For the reference technique obtained from RQ1, i.e., Harmonic Mean, voxel size did not produce a statistically significant global effect, although 0.02 achieved the highest macro ROC-AUC and provided the best balance between performance, coverage, score discrimination, and computational cost. Some techniques were substantially more sensitive to spatial granularity: Tarantula and Tarwid were strongly favored by coarse voxels, whereas Zoltar and Mountford benefited from finer voxels. Ochiai2 also deteriorated significantly at the finest resolution relative to Harmonic Mean. Thus, voxel size had to be fixed or explicitly controlled when comparing SBFL techniques.
\end{custombox}

\subsection{RQ4 -- Voxel Coverage Sensitivity}


\begin{table*}[t]
\centering
\caption{RQ4 learning-curve coverage and held-out performance by VLA. Coverage values are percentages. The Harmonic Mean cell reports ROC-AUC and its within-VLA, within-size rank among the 34 techniques in parentheses; leaders were determined from unrounded ROC-AUC.}
\label{tab:RQ4v2-coverage-performance}
{\scriptsize
\setlength{\tabcolsep}{4pt}
\begin{tabular}{lrrrrlr}
\toprule
VLA & $n$ & Calibration cov. (\%) & Held-out cov. (\%) &
Harmonic Mean AUC (rank) & Leading technique & Leading AUC \\
\midrule
EO1 & 10 & 8.2 & 8.2 & 0.5608 (10) & Ochiai2 & 0.5684 \\
 & 20 & 15.3 & 15.3 & 0.5688 (16) & Rogot1 / Scott & 0.5824 \\
 & 50 & 31.7 & 31.7 & 0.5938 (9) & Dennis & 0.6029 \\
 & 100 & 49.3 & 49.3 & 0.6174 (3) & Dennis & 0.6199 \\
 & 200 & 67.6 & 67.6 & 0.6377 (2.5) & Zoltar & 0.6381 \\
\midrule
GR00T & 10 & 9.1 & 9.1 & 0.5213 (7) & Dennis & 0.5226 \\
 & 20 & 16.8 & 16.8 & 0.5350 (4) & Dennis & 0.5375 \\
 & 50 & 34.1 & 34.1 & 0.5656 (2) & Dennis & 0.5667 \\
 & 100 & 51.5 & 51.6 & 0.5902 (3) & Zoltar & 0.5906 \\
 & 200 & 68.6 & 68.6 & 0.6217 (3.5) & Mountford & 0.6220 \\
\midrule
Pi0 & 10 & 5.1 & 5.1 & 0.5764 (5) & Rogot2 & 0.5799 \\
 & 20 & 9.6 & 9.7 & 0.6043 (6) & Arithmetic Mean & 0.6064 \\
 & 50 & 21.5 & 21.6 & 0.6225 (16) & Ochiai2 & 0.6389 \\
 & 100 & 36.5 & 36.6 & 0.6437 (15.5) & Arithmetic Mean & 0.6586 \\
 & 200 & 55.6 & 55.7 & 0.6707 (11.5) & Phi (Geometric Mean) & 0.6826 \\
\midrule
SpatialVLA & 10 & 6.0 & 6.0 & 0.5403 (8) &
Tarantula / Tarwid & 0.5482 \\
 & 20 & 11.2 & 11.2 & 0.5654 (5) & Zoltar & 0.5709 \\
 & 50 & 23.9 & 23.9 & 0.6101 (3) & Zoltar & 0.6151 \\
 & 100 & 39.0 & 39.0 & 0.6480 (3.5) & Zoltar & 0.6502 \\
 & 200 & 57.0 & 57.0 & 0.6833 (3.5) & Mountford & 0.6840 \\
\bottomrule
\end{tabular}
}
\end{table*}

\begin{table*}[t]
\centering
\caption{RQ4 sample efficiency for techniques reaching a top-five NAULC rank for at least one VLA. VLA and macro cells report NAULC with rank among the 34 techniques in parentheses. Bold denotes rank 1 and underlining the remaining top-five ranks. Endpoint and gain are macro ROC-AUC.}
\label{tab:RQ4v2-sample-efficiency}
{\scriptsize
\setlength{\tabcolsep}{3.5pt}
\begin{tabular}{lcccccrr}
\toprule
Technique & EO1 & GR00T & Pi0 & SpatialVLA & Macro & AUC@200 &
$\Delta$AUC$_{10\rightarrow200}$ \\
\midrule
Dennis
 & \textbf{0.6000 (1)}
 & \textbf{0.5646 (1)}
 & \underline{0.6302 (3)}
 & \underline{0.6026 (5)}
 & \textbf{0.5994 (1)}
 & 0.6501 & +0.0958 \\
Rogot2
 & \underline{0.5941 (2)}
 & \underline{0.5644 (3)}
 & 0.6232 (15)
 & \underline{0.6072 (4)}
 & \underline{0.5972 (2)}
 & 0.6534 & +0.1024 \\
Harmonic Mean
 & \underline{0.5938 (3)}
 & \underline{0.5645 (2)}
 & 0.6227 (16)
 & \underline{0.6072 (3)}
 & \underline{0.5971 (3)}
 & 0.6534 & +0.1036 \\
Phi (Geometric Mean)
 & \underline{0.5929 (4)}
 & \underline{0.5632 (5)}
 & \underline{0.6310 (2)}
 & 0.5956 (6)
 & \underline{0.5957 (4)}
 & 0.6466 & +0.0969 \\
Mountford
 & 0.5910 (9)
 & \underline{0.5638 (4)}
 & 0.6130 (17)
 & \underline{0.6092 (2)}
 & \underline{0.5942 (5)}
 & 0.6501 & +0.1032 \\
Arithmetic Mean
 & 0.5909 (10)
 & 0.5585 (7)
 & \textbf{0.6317 (1)}
 & 0.5795 (7)
 & 0.5902 (6)
 & 0.6294 & +0.0796 \\
Zoltar
 & 0.5816 (18)
 & 0.5621 (6)
 & 0.6048 (18)
 & \textbf{0.6108 (1)}
 & 0.5898 (7)
 & 0.6484 & +0.1150 \\
Ochiai2
 & \underline{0.5922 (5)}
 & 0.5445 (15)
 & \underline{0.6302 (4)}
 & 0.5325 (18)
 & 0.5748 (15)
 & 0.5882 & +0.0381 \\
\bottomrule
\end{tabular}
}
\end{table*}

\begin{table}[t]
\centering
\caption{RQ4 matched adjacent-size effects for Harmonic Mean over the 16 VLA--trajectory-generation blocks. Tests used paired Wilcoxon signed-rank tests with one Holm correction.}
\label{tab:RQ4v2-adjacent-effects}
{\scriptsize
\setlength{\tabcolsep}{4pt}
\begin{tabular}{lrrrrr}
\toprule
Transition & Mean $\Delta$AUC & Median $\Delta$AUC & Positive (\%) & Holm $p$ & $r_{rb}$ \\
\midrule
10--20 & +0.0187 & +0.0170 & 100.0 & 0.00012 & +1.000 \\
20--50 & +0.0296 & +0.0288 & 100.0 & 0.00012 & +1.000 \\
50--100 & +0.0268 & +0.0245 & 100.0 & 0.00012 & +1.000 \\
100--200 & +0.0285 & +0.0299 & 100.0 & 0.00012 & +1.000 \\
\bottomrule
\end{tabular}
}
\end{table}

Table~\ref{tab:RQ4v2-coverage-performance} reports the voxel coverage and held-out performance obtained as the number of spectrum-construction trajectories increased. Calibration coverage increased from 8.2\% to 67.6\% for EO1, from 9.1\% to 68.6\% for GR00T, from 5.1\% to 55.6\% for Pi0, and from 6.0\% to 57.0\% for SpatialVLA. Calibration and held-out coverage remained almost identical throughout the experiment, with a maximum difference of only 0.071 percentage points. This suggested that the calibration runs provided a representative estimate of the spatial coverage subsequently observed in the untouched test runs.

Harmonic Mean's ROC-AUC increased with every training-set enlargement for all four VLA models. Between 10 and 200 trajectories, its ROC-AUC increased from 0.5608 to 0.6377 for EO1, from 0.5213 to 0.6217 for GR00T, from 0.5764 to 0.6707 for Pi0, and from 0.5403 to 0.6833 for SpatialVLA. The corresponding improvements were 0.0769, 0.1004, 0.0942, and 0.1430, respectively. SpatialVLA therefore benefited most from the additional trajectories, whereas EO1 exhibited the smallest improvement. Nevertheless, the positive trend was observed consistently across all four models.

The descriptively leading technique also changed with both the VLA model and the number of trajectories. For EO1, Ochiai2 led at 10 trajectories, Rogot1 and Scott tied at 20, Dennis led at 50 and 100, and Zoltar led at 200. Dennis led the GR00T configurations with 10, 20, and 50 trajectories, whereas Zoltar led at 100 and Mountford at 200. For Pi0, Rogot2 led at 10 trajectories, Arithmetic Mean at 20, Ochiai2 at 50, Arithmetic Mean again at 100, and Phi at 200. For SpatialVLA, Tarantula and Tarwid tied at 10 trajectories, Zoltar led between 20 and 100, and Mountford led at 200. These changes indicated that the most effective suspiciousness formula depended not only on the VLA model but also on the amount of data available for constructing its spectra.


Table~\ref{tab:RQ4v2-sample-efficiency} summarizes performance over the complete learning curve using the normalized area under the learning curve (NAULC). Dennis obtained the highest macro NAULC, with 0.5994, followed closely by Rogot2 with 0.5972 and Harmonic Mean with 0.5971. Dennis's advantage over Harmonic Mean was therefore minimal. Moreover, Dennis did not lead every VLA: it ranked first for EO1 and GR00T, Arithmetic Mean ranked first for Pi0, and Zoltar ranked first for SpatialVLA. Harmonic Mean remained competitive for EO1, GR00T, and SpatialVLA but ranked only 16th for Pi0. Ochiai2 also showed substantial model dependence: it ranked fifth for EO1 and fourth for Pi0, but only 15th for GR00T and 18th for SpatialVLA, resulting in a macro rank of 15. Consequently, no technique dominated the complete learning curve for every VLA model.

The endpoint results differed slightly from the sample-efficiency ranking. At 200 trajectories, Harmonic Mean and Rogot2 jointly obtained the highest macro ROC-AUC of 0.6534, compared with 0.6501 for Dennis. Zoltar exhibited the largest improvement between 10 and 200 trajectories, increasing by 0.1150, but its lower performance with small training sets reduced its macro NAULC to 0.5898. These results illustrated the distinction between selecting a technique for strong performance with limited data and selecting one for its final performance after a larger spectrum-construction suite had become available.

Table~\ref{tab:RQ4v2-adjacent-effects} presents the matched statistical comparisons for Harmonic Mean. Every increase in training-set size produced a positive mean ROC-AUC change: 0.0187 from 10 to 20 trajectories, 0.0296 from 20 to 50, 0.0268 from 50 to 100, and 0.0285 from 100 to 200. Each improvement was positive in all 16 matched VLA--trajectory-generation blocks. All four comparisons remained statistically significant after Holm correction ($p_{\mathrm{Holm}}=0.00012$) and obtained a rank-biserial effect size of \(1.0\). The evidence therefore showed that the performance improvements were consistently observed across all evaluated VLA models and trajectory-generation conditions, rather than being driven by only a few particularly favorable experimental blocks.


Finally, Table~\ref{tab:RQ4v2-conditional-coverage} examines whether coverage remained associated with ROC-AUC after controlling for training-set size, configuration, and outer fold. The estimated change per additional ten percentage points of coverage ranged from \(-0.0088\) for Baroni-Urbani \& Buser to \(+0.0047\) for Zoltar. None of the 34 techniques exhibited a statistically significant conditional relationship after Holm correction; all adjusted \(p\)-values were \(1.000\). Thus, although coverage and ROC-AUC generally increased together as more trajectories were added, the results did not provide evidence that coverage independently explained the performance improvement. A plausible explanation was that additional trajectories improved the spectrum counts for already-observed voxels while also discovering new voxels, whereas the scalar coverage measure captured only the latter effect.

\begin{table*}[t]
\centering
\caption{Complete conditional coverage analysis for RQ4. Effects are ROC-AUC changes per ten percentage points of calibration coverage after centering within configuration, outer fold, and suite size. Holm correction covered all 34 techniques.}
\label{tab:RQ4v2-conditional-coverage}
{\scriptsize
\setlength{\tabcolsep}{6pt}
\begin{tabular}{lrrrrr}
\toprule
Technique & Mean effect & Median effect & Positive blocks (\%) & Holm $p$ & $r_{rb}$ \\
\midrule
Zoltar & +0.0047 & +0.0063 & 50.0 & 1.0000 & +0.206 \\
Gower & +0.0017 & +0.0034 & 56.2 & 1.0000 & +0.118 \\
Mountford & +0.0009 & +0.0046 & 56.2 & 1.0000 & -0.015 \\
Braun-Banquet & +0.0006 & +0.0035 & 62.5 & 1.0000 & +0.118 \\
Dennis & +0.0004 & -0.0014 & 43.8 & 1.0000 & +0.029 \\
Cohen & -0.0012 & -0.0053 & 37.5 & 1.0000 & -0.074 \\
Hamann & -0.0012 & -0.0053 & 37.5 & 1.0000 & -0.074 \\
Hamming & -0.0012 & -0.0053 & 37.5 & 1.0000 & -0.074 \\
Michael & -0.0012 & -0.0053 & 37.5 & 1.0000 & -0.074 \\
Rogers \& Tanimoto & -0.0012 & -0.0053 & 37.5 & 1.0000 & -0.074 \\
Simple Matching & -0.0012 & -0.0053 & 37.5 & 1.0000 & -0.074 \\
Sokal & -0.0012 & -0.0053 & 37.5 & 1.0000 & -0.074 \\
Rogot2 & -0.0012 & +0.0031 & 50.0 & 1.0000 & -0.147 \\
Harmonic Mean & -0.0014 & +0.0025 & 50.0 & 1.0000 & -0.162 \\
Phi (Geometric Mean) & -0.0014 & +0.0000 & 50.0 & 1.0000 & -0.044 \\
Arithmetic Mean & -0.0015 & -0.0014 & 43.8 & 1.0000 & -0.088 \\
Rogot1 & -0.0021 & +0.0001 & 50.0 & 1.0000 & -0.147 \\
Scott & -0.0021 & +0.0000 & 50.0 & 1.0000 & -0.147 \\
Anderberg & -0.0022 & -0.0010 & 43.8 & 1.0000 & -0.162 \\
Dice & -0.0022 & -0.0010 & 43.8 & 1.0000 & -0.162 \\
Goodman & -0.0022 & -0.0010 & 43.8 & 1.0000 & -0.162 \\
Jaccard & -0.0022 & -0.0010 & 43.8 & 1.0000 & -0.162 \\
Kulczynski & -0.0022 & -0.0010 & 43.8 & 1.0000 & -0.162 \\
Sorensen-Dice & -0.0022 & -0.0010 & 43.8 & 1.0000 & -0.162 \\
Fleiss & -0.0028 & -0.0009 & 50.0 & 1.0000 & -0.176 \\
Pearson & -0.0030 & -0.0064 & 31.2 & 1.0000 & -0.176 \\
Fossum & -0.0032 & +0.0009 & 50.0 & 1.0000 & -0.162 \\
Ochiai2 & -0.0032 & -0.0043 & 43.8 & 1.0000 & -0.176 \\
Ample & -0.0034 & -0.0032 & 50.0 & 1.0000 & -0.147 \\
Pierce & -0.0041 & -0.0040 & 43.8 & 1.0000 & -0.338 \\
Ochiai & -0.0045 & -0.0003 & 50.0 & 1.0000 & -0.162 \\
Tarwid & -0.0068 & -0.0067 & 37.5 & 1.0000 & -0.441 \\
Tarantula & -0.0068 & -0.0067 & 37.5 & 1.0000 & -0.441 \\
Baroni-Urbani \& Buser & -0.0088 & -0.0034 & 50.0 & 1.0000 & -0.294 \\
\bottomrule
\end{tabular}
}
\end{table*}

Overall, increasing the number of trajectories consistently enlarged voxel coverage and improved the failure discrimination of Harmonic Mean across all evaluated VLA models and trajectory-generation conditions. However, the conditional analysis did not show that coverage alone explained these improvements, suggesting that the additional spectrum evidence provided by larger training sets also played an important role. Moreover, the technique that used this additional information most effectively varied across VLA models and training-set sizes.

\begin{custombox}{Answer to RQ4}
Increasing the spectrum-construction suite from 10 to 200 trajectories consistently increased voxel coverage and significantly improved the ROC-AUC of Harmonic Mean across all evaluated VLA models and trajectory-generation conditions. Nevertheless, after controlling for training-set size, the analysis found no statistically significant independent association between greater coverage and performance for any of the 34 techniques. Dennis achieved the highest observed macro NAULC, whereas Harmonic Mean and Rogot2 jointly achieved the highest macro ROC-AUC at 200 trajectories. The best-performing technique therefore depended on both the VLA model and the number of available trajectories.
\end{custombox}

\subsection{RQ5: Cross-Task Generalizability}
To assess cross-task generalizability, we constructed an evaluation subset for each VLA model from the 504 executions performed on the Move-Near task. This resulted in 298 selected executions for Pi0, 246 for GR00T, 223 for EO1, and 342 for SpatialVLA. Executions were randomly selected using a fixed seed, with outcome-stratified sampling and spatial constraints to approximately preserve the failure proportion observed in RQ1. The selected trajectories were evaluated using the suspiciousness maps reconstructed from the Pick-Up training data, without retraining or calibration on Move-Near. This setup examined whether failure-associated information captured from Pick-Up trajectories remained useful for distinguishing passing from failing executions on a different task. 

Table~\ref{tab:rq5-transfer-roc-auc} reports the ROC-AUC values of the 34 SBFL techniques for each VLA model and their overall macro-average across the four models. Zoltar achieved the highest overall value, at 0.589, followed by Mountford at 0.572. Harmonic Mean and Rogot2, the RQ1 reference techniques, tied for third place, also with an overall value of 0.572 at the reported precision. Gower completed the top five overall ranks with a value of 0.561. Therefore, although RQ1 reference techniques remained among the strongest overall performers, their observed discrimination performance was lower on the Move-Near task than on the Pick-Up. 

\begin{table*}[t]
\centering
\caption{RQ5 -- Macro ROC-AUC of the 34 SBFL techniques for each VLA model on the Move-Near task. Parentheses report tie-aware ranks. Bold denotes rank 1 and underlining denotes the remaining top-five ranks. Harmonic Mean and Rogot2 are tied because they produce identical results in every reported column.}
\label{tab:rq5-transfer-roc-auc}
{\scriptsize
\setlength{\tabcolsep}{4pt}
\renewcommand{\arraystretch}{0.96}
\begin{tabular}{lccccc}
\toprule
Technique & EO1 & GR00T & Pi0 & SpatialVLA & Overall \\
\midrule
Zoltar & 0.604 (15) & \underline{0.571 (4)} & \textbf{0.677 (1)} & \underline{0.504 (3)} & \textbf{0.589 (1)} \\
Mountford & 0.593 (20) & \underline{0.568 (5)} & \underline{0.632 (2)} & 0.494 (7) & \underline{0.572 (2)} \\
Harmonic Mean & 0.618 (6) & \underline{0.574 (2)} & \underline{0.585 (4)} & \textbf{0.511 (1)} & \underline{0.572 (3)} \\
Rogot2 & 0.618 (6) & \underline{0.574 (2)} & \underline{0.585 (4)} & \textbf{0.511 (1)} & \underline{0.572 (3)} \\
Gower & 0.535 (33) & \textbf{0.605 (1)} & \underline{0.606 (3)} & 0.497 (6) & \underline{0.561 (5)} \\
Dennis & \underline{0.650 (2)} & 0.537 (9) & 0.521 (9) & \underline{0.502 (4)} & 0.552 (6) \\
Phi (Geometric Mean) & \textbf{0.650 (1)} & 0.538 (8) & 0.521 (8) & \underline{0.499 (5)} & 0.552 (7) \\
Tarwid & 0.608 (13) & 0.551 (6) & 0.525 (7) & 0.493 (9) & 0.544 (8) \\
Tarantula & 0.608 (14) & 0.551 (7) & 0.525 (6) & 0.493 (8) & 0.544 (9) \\
Arithmetic Mean & \underline{0.645 (3)} & 0.533 (13) & 0.513 (12) & 0.480 (10) & 0.543 (10) \\
Michael & \underline{0.643 (4)} & 0.533 (12) & 0.514 (11) & 0.473 (12) & 0.541 (11) \\
Cohen & \underline{0.639 (5)} & 0.533 (11) & 0.514 (10) & 0.472 (13) & 0.540 (12) \\
Hamann & 0.612 (8) & 0.517 (15) & 0.493 (13) & 0.452 (29) & 0.519 (13) \\
Hamming & 0.612 (8) & 0.517 (15) & 0.493 (13) & 0.452 (29) & 0.519 (13) \\
Rogers \& Tanimoto & 0.612 (8) & 0.517 (15) & 0.493 (13) & 0.452 (29) & 0.519 (13) \\
Simple Matching & 0.612 (8) & 0.517 (15) & 0.493 (13) & 0.452 (29) & 0.519 (13) \\
Sokal & 0.612 (8) & 0.517 (15) & 0.493 (13) & 0.452 (29) & 0.519 (13) \\
Ochiai2 & 0.604 (16) & 0.511 (20) & 0.492 (18) & 0.452 (28) & 0.515 (18) \\
Fleiss & 0.604 (17) & 0.501 (23) & 0.484 (19) & 0.452 (25) & 0.510 (19) \\
Ochiai & 0.592 (21) & 0.504 (21) & 0.482 (20) & 0.453 (15) & 0.508 (20) \\
Fossum & 0.591 (22) & 0.504 (22) & 0.481 (21) & 0.453 (16) & 0.507 (21) \\
Rogot1 & 0.599 (18) & 0.495 (24) & 0.479 (22) & 0.452 (26) & 0.506 (22) \\
Scott & 0.599 (18) & 0.495 (24) & 0.479 (22) & 0.452 (26) & 0.506 (22) \\
Baroni-Urbani \& Buser & 0.580 (23) & 0.487 (26) & 0.473 (30) & 0.453 (24) & 0.498 (24) \\
Anderberg & 0.576 (25) & 0.486 (27) & 0.473 (24) & 0.453 (17) & 0.497 (25) \\
Dice & 0.576 (25) & 0.486 (27) & 0.473 (24) & 0.453 (17) & 0.497 (25) \\
Goodman & 0.576 (25) & 0.486 (27) & 0.473 (24) & 0.453 (17) & 0.497 (25) \\
Jaccard & 0.576 (25) & 0.486 (27) & 0.473 (24) & 0.453 (17) & 0.497 (25) \\
Kulczynski & 0.576 (25) & 0.486 (27) & 0.473 (24) & 0.453 (17) & 0.497 (25) \\
Sorensen-Dice & 0.576 (25) & 0.486 (27) & 0.473 (24) & 0.453 (17) & 0.497 (25) \\
Braun-Banquet & 0.574 (31) & 0.485 (33) & 0.473 (31) & 0.453 (23) & 0.496 (31) \\
Pierce & 0.546 (32) & 0.470 (34) & 0.453 (32) & 0.451 (34) & 0.480 (32) \\
Ample & 0.576 (24) & 0.521 (14) & 0.344 (33) & 0.477 (11) & 0.479 (33) \\
Pearson & 0.513 (34) & 0.535 (10) & 0.260 (34) & 0.456 (14) & 0.441 (34) \\
\bottomrule
\end{tabular}
}
\end{table*}

Unlike in RQ1, where Harmonic Mean and Rogot2 ranked first for all four VLA models, no technique led across all models on the Move-Near task. Phi (Geometric Mean) ranked first for EO1, followed by Dennis. Gower ranked first for GR00T, while Zoltar led for Pi0. Harmonic Mean and Rogot2 shared first place for SpatialVLA. These results indicate that the strongest transferred technique depended on the VLA model. The model-specific rankings also showed why the overall results should be interpreted alongside the individual columns. Although Zoltar ranked first overall, it was the fifteenth for EO1, fourth for GR00T, first for Pi0, and third for SpatialVLA. Mountford ranked second for Pi0 and fifth for GR00T, but twentieth for EO1 and seventh for SpatialVLA. Harmonic Mean and Rogot2 were more consistent, as they ranked second for GR00T, fourth for Pi0, and first for SpatialVLA, although they placed sixth for EO1. The degree of discrimination also varied across models. The best transferred techniques achieved ROC-AUC values of 0.650 for EO1, 0.605 for GR00T, and 0.677 for Pi0. In contrast, the highest value for SpatialVLA was 0.511, indicating discrimination close to chance level (i.e., random discrimination). Several techniques also produced values below 0.5 for individual models. For example, Pearson obtained 0.260 for Pi0, while Ample obtained 0.344. Therefore, transferring a suspiciousness map did not consistently preserve useful failure discrimination across techniques and models. Several groups of techniques shared the same reported values and ranks. Harmonic Mean and Rogot2 tied in every column. The same pattern occurred for Hamann, Hamming, Rogers \& Tanimoto, Simple Matching, and Sokal; and for Anderberg, Dice, Goodman, Jaccard, Kulczynski, and Sorensen-Dice. These similarities are consistent with mathematical relationships among the formulas or equivalent trajectory rankings on the evaluated data.

Compared with RQ1, Harmonic Mean and Rogot2 decreased from an overall macro ROC-AUC of 0.704 to approximately 0.572, a reduction of 0.132. The largest decrease occurred for SpatialVLA, from 0.732 to 0.511, followed by Pi0, from 0.746 to 0.585. Their values also decreased for GR00T, from 0.673 to 0.574, and for EO1, from 0.665 to 0.618. Thus, the discrimination achieved within the Pick-Up task transferred only partially, with the extent of the decrease depending on the model. One plausible explanation is that regions associated with failures during Pick-Up are less informative about failures during the Move-Near task. Moreover, the ordering of the techniques also changed between tasks. Zoltar moved from tenth overall in RQ1 to first on Move-Near, despite its overall macro ROC-AUC decreasing from 0.643 to 0.589. Mountford moved from twelfth to second overall, while its value decreased from 0.586 to 0.572. These changes show that an improvement in relative rank does not necessarily imply an improvement in discrimination. They also indicate that the technique rankings observed in RQ1 did not remain stable after transfer.

When comparing against the baselines, as shown in Table~\ref{tab:rq5-baseline-performance}, among the generic baselines, TCP Acceleration Instability achieved the highest overall macro ROC-AUC, at 0.6172, followed by Trajectory Instability at 0.6162 and TCP Velocity Instability at 0.5985. All three exceeded Zoltar, the strongest transferred SBFL technique overall, at 0.5892. Harmonic Mean and Rogot2 each achieved 0.5718. Consequently, the strongest generic baselines provided better overall discrimination than the transferred suspiciousness maps on the evaluated samples. Nevertheless, this comparison differed across models. For EO1, TCP Acceleration Instability achieved 0.7881, exceeding the best transferred SBFL result of 0.650 from Phi (Geometric Mean). For GR00T, Action Position Instability achieved 0.6568, compared with 0.6047 for Gower, the strongest transferred SBFL technique for that model. For Pi0, Zoltar achieved 0.6771, slightly exceeding the strongest generic baseline, TCP Acceleration Instability, at 0.6708. For SpatialVLA, Harmonic Mean and Rogot2 achieved 0.5105, exceeding the best generic baseline, TCP Acceleration Instability, at 0.4682. However, the SpatialVLA results remained close to chance level even for the leading transferred techniques. Thus, generic baselines performed better for EO1 and GR00T, whereas transferred SBFL achieved higher values for Pi0 and SpatialVLA. These results suggest that, although the baselines performed better overall, metric selection should account for both the VLA model and the target task.

\begin{table*}[ht]
\centering
\caption{RQ5 baseline trajectory-level failure-discrimination ROC-AUC on the selected RQ5 evaluation samples. Overall is the equal-weight mean across the four models. Bold denotes the best value and underlining the second-best value per column. $^{\dagger}$ indicates a task-aware metric using object/goal state and $^{\star}$ indicates an SBFL technique.}
\label{tab:rq5-baseline-performance}
{\scriptsize
\setlength{\tabcolsep}{4pt}
\begin{tabular}{lcccccc}
\toprule
Method & EO1 & GR00T & Pi0 & SpatialVLA & Overall & Rank \\
\midrule
Optimal Trajectory$^{\dagger}$ & \textbf{0.8980} & \textbf{0.8742} & \textbf{0.9473} & \textbf{0.7644} & \textbf{0.8710} & 1 \\
TCP Acceleration Instability & \underline{0.7881} & 0.5416 & 0.6708 & 0.4682 & \underline{0.6172} & 2 \\
Trajectory Instability & 0.7858 & 0.5529 & 0.6615 & 0.4645 & 0.6162 & 3 \\
TCP Velocity Instability & 0.7752 & 0.5618 & 0.6285 & 0.4286 & 0.5985 & 4 \\
Zoltar$^{\star}$ & 0.6041 & 0.5714 & \underline{0.6771} & 0.5044 & 0.5892 & 5 \\
Action Position Instability & 0.7724 & \underline{0.6568} & 0.5065 & 0.4125 & 0.5871 & 6 \\
Action Velocity Instability & 0.7723 & 0.6441 & 0.5011 & 0.4131 & 0.5826 & 7 \\
Action Acceleration Instability & 0.7707 & 0.6381 & 0.4920 & 0.4027 & 0.5759 & 8 \\
Mountford$^{\star}$ & 0.5926 & 0.5682 & 0.6320 & 0.4945 & 0.5718 & 9 \\
Harmonic Mean$^{\star}$ & 0.6182 & 0.5738 & 0.5846 & \underline{0.5105} & 0.5718 & 10 \\
Rogot2$^{\star}$ & 0.6182 & 0.5738 & 0.5846 & \underline{0.5105} & 0.5718 & 10 \\
TCP Position Instability & 0.7285 & 0.6412 & 0.5816 & 0.3317 & 0.5708 & 12 \\
Gower$^{\star}$ & 0.5347 & 0.6047 & 0.6064 & 0.4973 & 0.5608 & 13 \\
\bottomrule
\end{tabular}
}
\end{table*}

Optimal Trajectory achieved the highest ROC-AUC for every VLA model. However, it uses task-specific object and goal information, whereas the generic baselines use trajectory characteristics, and the transferred SBFL maps use failure-associated information from Pick-Up. These differences in the information required by each method should be considered alongside their discrimination performance when selecting a metric for a target task.

\begin{custombox}{Answer to RQ5}
Under the studied conditions, suspiciousness maps constructed from Pick-Up trajectories retained some ability to distinguish passing from failing Move-Near executions, but their effectiveness depended on the VLA model and SBFL technique. The strongest generic baselines outperformed transferred SBFL overall and for EO1 and GR00T, while transferred SBFL achieved higher values for Pi0 and SpatialVLA. Therefore, transferred suspiciousness maps should be validated on the intended target task before being relied upon in operation.
\end{custombox}

\subsection{Concluding Remark and Discussion}

The results show that the use of VoxelSBFL provides useful evidence for distinguishing successful from failed VLA executions. However, its effectiveness depends on the policy, task, and spectrum-construction setting. Although Harmonic Mean and Rogot2 led at the reference voxel size, changes in granularity, training-set size, and task affected technique rankings. These findings inform the potential applications of VoxelSBFL and the conditions under which its maps can be reused.

\subsubsection{Key findings}

Our study leverages four key findings that are worth discussing:

\textbf{Finding 1 -- SBFL can distinguish successful and failed VLA executions from physical workspace coverage:} 
By treating workspace voxels as coverage elements, VoxelSBFL applies spectrum-based analysis to the observable behavior of a robot. It constructs spectra from recorded tool-center-point trajectories and episode outcomes, which are available independently of the policy's architecture. At the reference voxel size, Harmonic Mean and Rogot2 ranked first for all four VLAs and achieved a macro ROC-AUC of 0.704. This result shows that \emph{physical workspace coverage can serve as a basis for applying SBFL to black-box VLA policies and that certain failures can be associated with some VLA workspace regions}.

\textbf{Finding 2 -- Voxel visitation reveals spatial patterns associated with failure:}
The discrimination achieved on held-out executions indicates that \emph{where the robot moves carries information about whether it succeeds}. VoxelSBFL makes this information inspectable by ranking workspace regions according to their association with failed executions. A highly ranked region may be where an unsuccessful grasp occurs, where the effects of an earlier error become visible, or where the robot attempts to recover. The map therefore helps narrow down which segments of a trajectory warrant inspection. Establishing the cause of an individual failure requires examining what the robot did in and around those segments.

\textbf{Finding 3 -- Spatial visitation and motion instability capture different aspects of failure:}
On the ``Pick-Up'' task, Harmonic Mean outperformed all seven action- and motion-instability baselines proposed by Valle et al.~\cite{valle2025evaluating} overall, while several motion-based metrics performed better for EO1. One plausible explanation is that a smooth trajectory can still take the end effector to an unsuitable location, while successful and failed trajectories that visit similar regions can differ in how the robot moves through them. The task-aware Optimal Trajectory metric achieved the highest overall discrimination using additional information about the object and goal. Together, these results motivate studying when spatial visitation, motion dynamics, and task context provide complementary evidence.

\begin{figure}[ht]
    \centering
    \includegraphics[width=\textwidth]
    {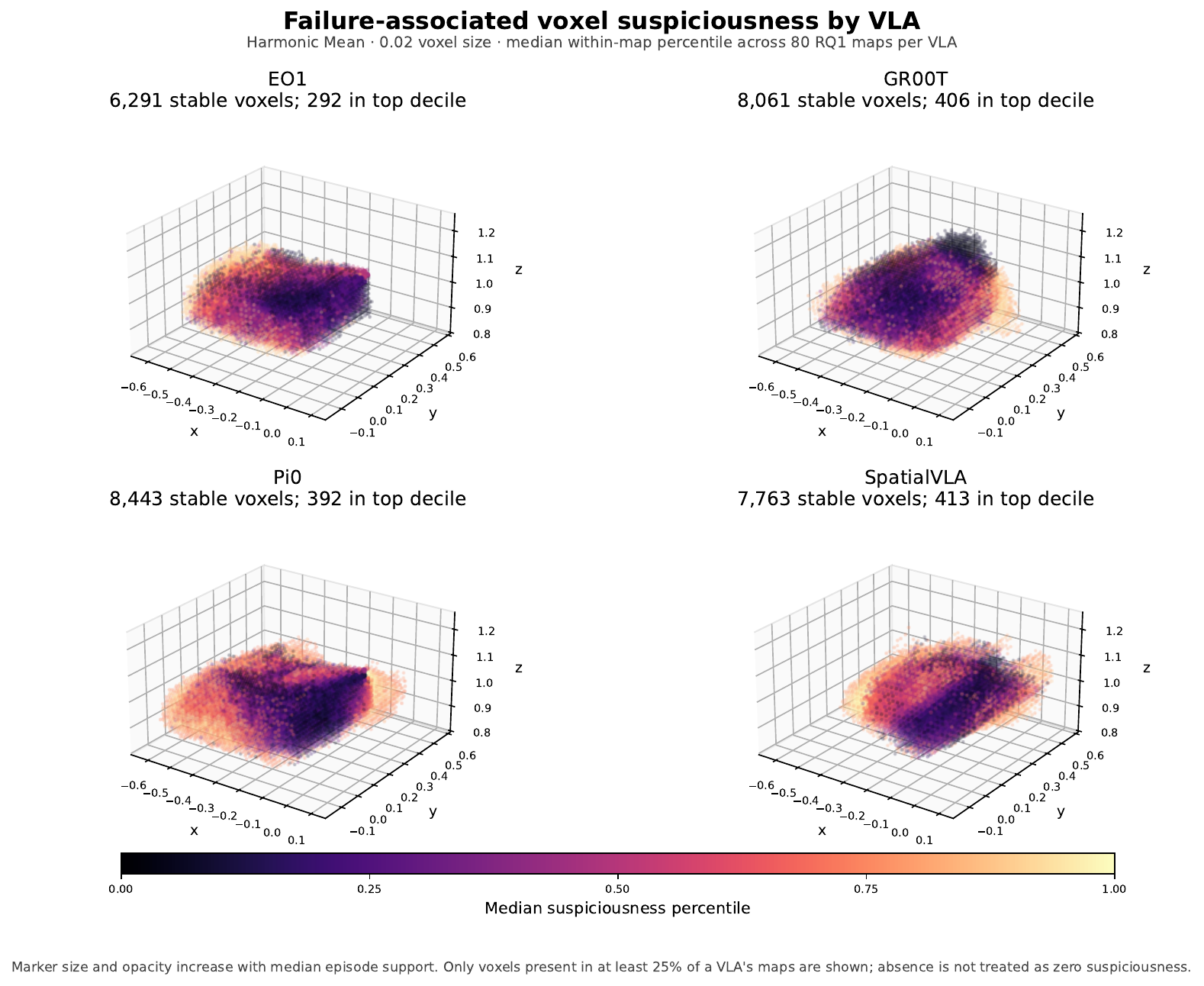}
    \caption{Three-dimensional distribution of failure-associated voxel
    suspiciousness for the four VLA models. Colors represent the median
    within-map percentile of the Harmonic Mean suspiciousness score at voxel
    size 0.02, computed across the 80 RQ1 maps available for each VLA.
    Warmer colors indicate higher suspiciousness, while marker size and opacity
    increase with episode support.}
    \label{fig:rq1-voxel-suspiciousness}
\end{figure}

\textbf{Finding 4 -- The learned spatial patterns depend on the policy and task:}
Figure~\ref{fig:rq1-voxel-suspiciousness} shows where the highest-ranked voxels occur for each VLA under the same Pick-Up task and robot embodiment. In the common viewing orientation, EO1's higher-ranked voxels appear mainly along the left flank of the visited volume, whereas GR00T shows a prominent higher-ranked patch on the right flank. Pi0 shows higher-ranked voxels on both outer flanks around a lower-ranked central region. For SpatialVLA, higher-ranked voxels extend along an elongated band beside a lower-ranked strip. These differences illustrate how the same SBFL formula can direct an engineer toward different parts of the workspace for different policies. Colors encode relative ranks within each policy's maps, and marker size and opacity indicate the amount of execution evidence supporting each voxel.

\subsubsection{Applicability and Implications: }
VoxelSBFL may have two distinct yet complementary applications:

\textbf{Runtime intervention and self-healing:} Voxel suspiciousness could serve as a signal for an external guardrail that supervises the VLA. When a trajectory enters a highly suspicious region, or is predicted to enter one, the guardrail could suspend VLA control, bring the robot to a controlled stop, and request human assistance. For a robot with a mobile base, the guardrail could instead transfer the main control to a navigation policy that repositions the robot relative to the target object. From the new base pose, the VLA could retry the manipulation from a different viewpoint and approach the object through a different part of its reachable workspace. This is a form of non-invasive runtime recovery in which the guardrail changes the conditions under which the VLA acts. Because our evaluation scores complete trajectories, a runtime study would need to establish whether alerts arrive early enough to intervene, how often they trigger unnecessary interventions, and whether repositioning improves task completion.

\textbf{Offline diagnosis and targeted VLA repair:} Engineers could use suspicious voxels to select segments of failed executions for inspection and compare them with successful executions under similar scene conditions. Reviewing camera observations, gripper actions, and object motion before and after a visit can help identify where the behaviors began to diverge. Engineers could then vary the object's initial pose to test whether the suspected behavior recurs and collect successful demonstrations under those conditions. For example, if inspection reveals that the gripper repeatedly closes before reaching the object, the new demonstrations could show the correct approach and closure timing. These demonstrations could support targeted fine-tuning, with follow-up tests assessing whether the revised policy succeeds in the scenarios that exposed the original failure.

\section{Threats to Validity}
\label{sec:ThreatstoValidity}

One threat to \textbf{construct validity} concerns the use of binary execution outcomes and ROC-AUC to evaluate failure discrimination. Although these measures allow us to distinguish successful from failed trajectories, they do not establish whether highly suspicious voxels correspond to the actual causes of failure. Furthermore, the evaluation considers complete trajectories rather than measuring failure anticipation during execution. To mitigate this threat, we applied consistent outcome labels and scoring procedures across the compared techniques and evaluated their discrimination capability on held-out trajectories. 


Regarding \textbf{internal validity}, the results may be influenced by the trajectory-generation strategies, spatial discretization, and selection of evaluation trajectories. In particular, the outcome-stratified sampling used in RQ5 may affect the observed transfer performance. To mitigate these threats, we evaluated the techniques on the same trajectories within each experimental configuration and separated training and evaluation through leave-one-run-out validation. Furthermore, we considered four trajectory-generation strategies and examined the influence of voxel size, training-set size, and execution coverage through sensitivity analyses in RQ3 and RQ4. For RQ5, spatial constraints were applied to ensure balanced coverage across the evaluation regions.

A potential threat to \textbf{conclusion validity} concerns the statistical interpretation of the experimental results. The experimental configurations share underlying factors, and the validation folds are not entirely independent, which may affect the interpretation of statistical significance. In addition, the reported confidence intervals are pointwise, and non-significant differences do not establish formal statistical equivalence. To mitigate these threats, we aggregated folds and scene budgets before performing paired statistical comparisons, applied exact Wilcoxon signed-rank tests with Holm correction, and reported effect sizes and confidence intervals where applicable. We also assessed the sensitivity of the bootstrap results and interpreted the cross-task differences in RQ5 descriptively. 

Finally, \textbf{external validity} may be affected by the selection of VLA models, tasks, and environments. Our evaluation was limited to four VLA models and two robotic manipulation tasks in simulation.
Consequently, the findings may not generalize to other VLA architectures, tasks, or environments. To mitigate this threat, we evaluated four different models from different architectural families across four trajectory-generation strategies and multiple experimental configurations, ensuring high diversity. We also examined cross-task transfer without retraining on the target task, providing an initial assessment of the generalizability beyond the original task. Another threat to external validity concerns the sim-to-real gap, as our experiments were conducted exclusively in simulation. Evaluating VoxelSBFL on a physical robot was not feasible because the required robotic platform was unavailable and reproducing thousands of trajectories would be impractical and costly. To partially mitigate this threat, we used SimplerEnv, whose authors reported a strong correlation between simulated and real-world performance of robotic manipulation policies~\cite{li24simpler}. 

\section{Related Work}
\label{sec:Relatedwork}

This section reviews prior work on testing and robustness evaluation of VLA models, failure prediction and uncertainty estimation in robotic policies, extensions of spectrum-based fault localization, and spatial representations for robotic manipulation. We position VoxelSBFL within this literature by highlighting its similarities to and differences from existing approaches.

\subsection{Failure Prediction and Uncertainty Estimation in Robotic Policies}
Failure prediction in learned robotic policies has been investigated using uncertainty metrics, execution-quality metrics, internal representations, and behavioral consistency. Valle et al.~\cite{valle2025evaluating} studied eight uncertainty metrics and five quality metrics for VLA-enabled robots, evaluating their agreement with human assessments and their ability to distinguish successful from unsuccessful executions. Their metrics characterize token distributions, action instability, variability across repeated inferences, and trajectory quality. Their information requirements differ since token-based metrics require prediction probabilities, action and trajectory metrics use policy--robot interface signals, and the Optimal Trajectory metric additionally uses object and goal information. Similar to this approach, confidence metrics have also been used for intervention and calibration. Karli et al.~\cite{karli2025ask} used token-level entropy and perplexity from a fine-tuned $\pi_0$-FAST policy to request human corrections and investigate uncertainty-weighted shared control. Zollo and Zemel~\cite{zollo2025confidence} examined confidence calibration over task execution and proposed prompt ensembles and action-wise Platt scaling. VoxelSBFL also assesses execution reliability, but learns spatial associations from TCP trajectories labeled with episode outcomes and scores unseen trajectories according to their visited regions. Its contribution is therefore a learned spatial representation of failure-associated behavior, whose percentiles express relative suspiciousness rather than calibrated failure probabilities. We directly compare this representation with three action-instability metrics and five quality metrics from Valle et al.~\cite{valle2025evaluating} that are used as baseline metrics.

Other methods exploit policy representations or temporal behavior. For example, Gu et al.~\cite{gu2026safe} proposed SAFE, learning a failure detector from internal VLA features and successful and failed rollouts, with conformal thresholds and evaluation on unseen tasks. Podolinsky et al.~\cite{podolinsky2026failure} introduced VLA-Tug, a recurrent probe trained on internal VLA representations with annotated failure onsets that predicts the likely failure mode and calibrates its alarm threshold to jointly control false alarms and missed early detections. Lyu et al.~\cite{lyu2026sentryvla} introduced SentryVLA, combining conformalized quantile regression for scoring stochastic action candidates with Mahalanobis-distance monitoring of latent execution states. Bhardwaj et al.~\cite{bhardwaj2026progress} introduced Progress Probes ($P^2$), which train linear probes on internal VLA activations from successful demonstrations to estimate task progress. Their runtime monitor flags executions whose estimated progress falls behind the nominal progression expected from successful executions. Agia et al.~\cite{agia2024unpacking} proposed Sentinel, combining Statistical Temporal Action Consistency with vision-language-model-based assessment of task progress. VoxelSBFL instead operates on externally recorded TCP positions, without requiring internal policy features, additional action sampling, or a vision-language monitor. Its map explicitly identifies physical regions associated with failed executions, although binary voxel coverage discards temporal ordering and visit frequency. These methods therefore provide methodological comparisons concerning representation and information requirements.

\subsection{Testing and Robustness Evaluation of Vision-Language-Action Models}

Existing studies investigate VLA reliability through simulation-based evaluation, systematic variation of execution conditions, and automated test generation. Li et al.~\cite{li24simpler} introduced SIMPLER, addressing visual and control discrepancies between real and simulated environments to support reproducible policy evaluation. Wang et al.~\cite{wang2025vlatest} proposed VLATest, a fuzzing framework that varies target objects, confounding objects, lighting, and camera pose, while also examining unseen objects and instruction mutations. Peng et al.~\cite{peng2025nebula} introduced NEBULA, combining capability tests with stress tests to characterize policy capabilities and robustness beyond aggregate task success. Other approaches address test oracles and failure discovery. For example, Valle et al.~\cite{valle2026metamorphic} proposed metamorphic relations that compare robot trajectories under related inputs, alleviating the need for an explicit correctness oracle for each execution. For VLA testing, Kanwal et al.~\cite{kanwal2026fatevla} proposed FATE-VLA, combining diversity-oriented exploration with surrogate predictions over scene configurations to discover diverse failures.

These approaches primarily support robustness evaluation, test input generation, and test-oracle design. Our work, in contrast, focuses on failure prediction by constructing spatial coverage spectra from robot trajectories and externally supplied episode outcomes. These spectra are used to compute suspiciousness scores for workspace voxels, identifying spatial regions associated with failed executions. The resulting suspiciousness map can then be used to anticipate failures and act on it by using specific guardrails (e.g., stopping the robotic movement and asking for human intervention). Beyond failure prediction, this map may also support failure analysis and explanation by highlighting regions to prioritize for further investigation and debugging, which may help future VLA repair techniques (e.g., by generating valid trajectories that target highly suspicious voxels and afterward retraining it). However, highly suspicious voxels indicate statistical associations with failure rather than establishing its underlying causes.

\subsection{Spectrum-Based Fault Localization and Its Extensions}

Several empirical studies have investigated the effectiveness of SBFL techniques under different experimental conditions. Jones and Harrold~\cite{jones2005empirical} compared Tarantula with four fault-localization techniques and reported higher localization effectiveness on their subject programs. Abreu et al.~\cite{abreu2009practical} investigated similarity coefficients and the influence of the quantity and quality of execution observations, finding favorable results for Ochiai. Pearson et al.~\cite{pearson2017evaluating} revisited comparisons between spectrum-based and mutation-based techniques using artificial and real faults. Their results showed that conclusions obtained with artificial faults did not necessarily hold for real faults, highlighting the importance of the evaluation context when comparing localization techniques. In addition, theoretical studies provide complementary perspectives on suspiciousness formulas. Naish et al.~\cite{naish2011model} analyzed spectral ranking methods using a model of software diagnosis, including relationships between equivalent rankings. Xie et al.~\cite{xie2013theoretical} investigated theoretical relationships between risk-evaluation formulas. Together with the broader survey by Wong et al.~\cite{wong2016survey}, these studies motivate examining established coefficients under different coverage representations rather than assuming that their relative performance transfers across contexts. 
These studies examine SBFL in the context of software fault localization. In contrast, our study adapts its coverage–outcome principle to the physical workspace of robotic manipulators. We recast the problem by treating workspace voxels as spectrum elements, TCP visitation as coverage, and episode success or failure as the execution outcome. This adaptation changes both the unit of analysis and the objective: instead of ranking potentially faulty code elements, we identify spatial regions associated with unsuccessful robotic executions and use their suspiciousness scores to predict failure in unseen trajectories.


SBFL has also been extended beyond conventional code-level bug localization. Arrieta et al.~\cite{arrieta2018spectrum} ranked software-product-line features and constructed small valid products containing suspicious features to support fault isolation. Eniser et al.~\cite{eniser2019deepfault} introduced DeepFault, which applies suspiciousness measures to neuron activation spectra and uses gradients to generate inputs exercising suspicious neurons. Sun et al.~\cite{sun2020explaining} proposed DeepCover, a black-box approach that ranks image features through statistical fault localization to explain classifier decisions. Addressing the oracle problem, Xie et al.~\cite{xie2013metamorphic} introduced metamorphic slices, combining program slicing with violations of metamorphic relations instead of requiring pass/fail outcomes for individual test cases. 
These extensions demonstrate the applicability of spectrum-based reasoning beyond conventional program elements. To the best of our knowledge, however, our work is the first to investigate SBFL in the context of VLA models. This domain introduces a different analysis target: the physical behavior produced by a learned robotic policy, observed through trajectories and episode outcomes. VoxelSBFL introduces a spatial spectrum representation that connects this behavior to execution failures, enabling established suspiciousness techniques (e.g., Tarantula) to identify failure-associated workspace regions and score unseen trajectories. Our study investigates the effectiveness of this formulation across VLA models and examines its sensitivity to spatial resolution, training coverage, and transfer between tasks.

\section{Conclusions and Future Work}
\label{sec:conclusion}

In this paper, we introduced \textit{VoxelSBFL}, a spatial adaptation of SBFL that allows for measuring suspiciousness levels of voxels in the robot's workspace. By replacing code elements with workspace voxels, the approach connects recorded TCP trajectories and episode outcomes with suspiciousness scores. The resulting maps identify failure-associated regions and score unseen trajectories without requiring internal model access or explicit object and goal positions. We evaluated 34 spectrum-based techniques across four VLA models using 96,000 Pick-Up trajectories and compared the leading techniques against eight baseline metrics from \cite{valle2025evaluating}. Harmonic Mean and Rogot2 achieved the highest observed performance at the reference voxel size, with an overall macro ROC-AUC of 0.704, and significantly outperformed the evaluated action- and motion-instability baselines overall. Nevertheless, Optimal Trajectory achieved stronger discrimination using additional task-specific information, although this specific metric has several limitations for real-world applications (i.e., it is task-specific and requires target object position information in operation, which is not available in most contexts). Our sensitivity analyses further showed that no voxel size was uniformly optimal across models and techniques. Increasing the training set size improved the reference technique's discrimination, although the analysis did not establish an independent relationship between spatial coverage and performance. Moreover, transfer to Move-Near yielded weaker discrimination, highlighting the task dependence of the learned spatial associations. These findings support spatial spectra as a useful representation for VLA execution assessment while emphasizing that suspicious regions indicate associations with failures.

Our work opens several research avenues. An interesting direction that requires a deep adaptation of our approach is to extend VoxelSBFL to robots with multiple arms, including humanoids. We will compare a shared workspace suspiciousness map with separate maps for each arm. We hypothesize that arm-specific maps will better capture failure-associated regions because the two arms may differ in reachability and dexterity even where their workspaces overlap. A related question is how to define voxel coverage when a single pass/fail outcome applies to the entire execution. For example, if one arm misses a grasp while the other remains idle, counting visits from both arms equally could assign high suspiciousness to the idle arm's position. We will investigate how to account for each arm's role at different stages of the task, such as grasping, stabilizing an object, or remaining idle. We will also examine whether tracking which voxels the arms occupy simultaneously helps identify failures that arise during coordinated actions such as handovers. A second direction is to use suspiciousness maps to guide targeted test generation and demonstration collection for policy improvement. This will involve prioritizing execution contexts associated with highly suspicious regions and assessing whether the additional demonstrations support effective fine-tuning. Finally, we plan to quantify the sim-to-real gap by evaluating whether suspiciousness maps learned in simulation can anticipate failures on physical robots. We will apply the same maps to partial trajectories from matched simulated and physical executions and measure how often they flag failures before onset at a fixed false-alarm rate, as well as how much lead time they provide. Comparing these results with those obtained using maps learned from physical executions will help determine how much of the gap can be closed with real-world data. 

\section*{Acknowledgments}
Arusa Kanwal, Miren Illarramendi, Shaukat Ali and Aitor Arrieta are supported by the InnoGuard Marie Skłodowska-Curie Doctoral Network (Grant Agreement No. 101169233). Pablo Valle is supported by the Pre-doctoral Program for the Formation of Non-Doctoral Research Staff of the Education Department of the Basque Government (Grant n. PRE\_2025\_2\_0252). Aitor Arrieta and Miren Illarramendi are partly supported by the Spanish Ministry of Science, Innovation and Universities (project PID2023-152979OA-I00), funded by MCIU /AEI /10.13039/ 501100011033/FEDER, UE. The authors are part of the Systems and Software Engineering group of Mondragon Unibertsitatea (IT1919-26), supported by the Department of Education, Universities and Research of the Basque Country. 

\bibliographystyle{ACM-Reference-Format}
\bibliography{sample-base}

\appendix

\end{document}